\documentclass{article} 
\usepackage{iclr2027_conference,times}

\usepackage{amsmath,amsfonts,bm}

\def\eqref#1{equation~\ref{#1}}

\def\1{\bm{1}}

\DeclareMathAlphabet{\mathsfit}{\encodingdefault}{\sfdefault}{m}{sl}
\SetMathAlphabet{\mathsfit}{bold}{\encodingdefault}{\sfdefault}{bx}{n}

\usepackage{booktabs}

\usepackage{graphicx}
\usepackage{wrapfig}

\usepackage{amsmath}
\usepackage{amssymb}
\usepackage{nicefrac}

\usepackage{multirow}
\usepackage{multicol}
\usepackage{algorithm}
\usepackage{algpseudocodex}

\usepackage[normalem]{ulem}

\usepackage{hyperref}
\usepackage[capitalize]{cleveref}
\crefname{algorithm}{Algorithm}{Algorithms}

\usepackage{url}
\usepackage[dvipsnames]{xcolor}     
\usepackage{colortbl}   
\definecolor{lightgray}{gray}{0.9}
\definecolor{midgray}{gray}{0.8}

\usepackage{threeparttable}

\usepackage{enumitem}

\definecolor{gaincolor}{RGB}{0,130,70}
  \definecolor{dropcolor}{RGB}{190,45,45}
  \definecolor{ourscolor}{RGB}{0,90,160}
  \definecolor{avggray}{RGB}{242,242,242}
  \definecolor{gainbg}{RGB}{228,246,235}

  \newcommand{\gain}[1]{\,{\scriptsize\color{gaincolor}(+#1)}}
  
  \newcommand{\ours}{\textbf{\color{ourscolor}Ours}}

\usepackage{amsthm}
\usepackage{subcaption}

\newtheorem{proposition}{Proposition}

\theoremstyle{remark}
\newtheorem*{proof*}{Proof}

\usepackage{changes}
\definechangesauthor[name={Cuong}, color=Red]{cn}

\title{Beyond Local Reconstruction: Geometry-Preserving Output Alignment for Binary LLM Quantization}

\author{Hoang Anh Dung \\
Department of Data Science and AI \\
Monash University\\
\texttt{\{hoang.dung\}@monash.edu} \\
\And
Cuong Pham \\
Department of Data Science and AI \\
Monash University\\
\texttt{\{cuong.pham\}@monash.edu} \\
\AND
Cuong Nguyen \\
Centre for Vision, Speech and Signal Processing \\
University of Surrey \\
\texttt{\{c.nguyen\}@surrey.ac.uk} \\
\AND
Trung Le \\
Department of Data Science and AI \\
Monash University\\
\texttt{\{trunglm\}@monash.edu} \\
{jianfei.cai\}@monash.edu} \\
\AND
Thanh-Toan Do \\
Department of Data Science and AI \\
Monash University\\
\texttt{\{toan.do\}@monash.edu} \\
\AND
Jianfei Cai \\
Department of Data Science and AI \\
Monash University\\
}

\iclrfinalcopy 
\begin{document}

\maketitle

\begin{abstract}
    Recent advances in post-training quantization for large language models (LLMs) have demonstrated that low-bit quantization methods can maintain most of the original model performance. Despite such successes, 1-bit quantization remains particularly challenging. A common strategy in 1-bit quantization is to determine binary weights by matching full-precision parameters, following a weight-driven criterion.  However, this objective is not directly aligned with the quantized model's objective, which is to preserve the model's output behavior under the impact of quantization. A natural alternative is to adopt output-driven criteria that minimize discrepancies in model outputs using calibration data. Surprisingly, naive output-driven approaches often perform even worse in the 1-bit regime.
    In this paper, we show that this failure arises from two fundamental issues: error accumulation across layers and, more critically, \emph{anisotropic distortion} of the representation space. 
    Based on this observation, we introduce a geometry-preserving output-alignment framework for 1-bit LLM quantization. First, we formulate output alignment as a \emph{constrained optimization} problem that improves reconstruction while explicitly controlling changes in token similarity, preventing reconstruction gains from being achieved at the expense of representation geometry. Second, we propose a \emph{geometry-aware salient-mask selection} strategy that identifies quantization configurations according to their ability to preserve token relations, rather than relying solely on reconstruction error.
    These components enable effective output-driven optimization while retaining the structural information lost by naive alignment. Extensive experiments across multiple LLM families and downstream tasks demonstrate consistent improvements over existing 1-bit PTQ methods.
\end{abstract}

\section{Introduction}

Large language models (LLMs)~\citep{Wei_2022,Radford2019LanguageMA,zhang2022opt,brown2020language} have become a focal point of both academic research and industrial development, thanks to their strong capabilities across a wide range of natural language processing tasks~\citep{Hendrycks_2020,bisk2020piqa}, including question answering \citep{Devlin2019BERTPO}, machine translation \citep{Fan2020BeyondEM,Lepikhin2020GShardSG}, summarization \citep{Zhang2019PEGASUSPW,Lewis2019BARTDS} and language generation \citep{Radford2019LanguageMA,brown2020language}. Despite these advances, the massive scale of modern LLMs, often involving billions of parameters, poses substantial challenges for efficient inference and deployment. To address this, the community has explored various compression approaches, such as 
knowledge distillation~\citep{hinton2015distilling}, network quantization~\citep{choi2018bridging,frantar2022gptq}, and pruning~\citep{han2015deep}. However, many of these approaches depend on large-scale training data and costly retraining, which limits their practicality. In contrast, post-training quantization (PTQ)~\citep{ding2023cbq,sun2024flatquant} requires only a small calibration set and modest computational resources, making it a practical choice for compressing LLMs. Despite strong empirical performance at near 4-bit quantization, the most extreme case, 1-bit quantization, remains highly challenging.

Existing 1-bit PTQ methods typically use a layer-wise framework with two stages. First, in the quantization stage, a block of weights is binarized by solving a closed-form objective for the quantized variables, while the remaining weights in the layer are kept in full precision. Second, in the error compensation stage, the residual quantization error is mitigated by updating these remaining full-precision weights. Prior approaches differ mainly in the objective used during the quantization stage: (1) \emph{weight-driven methods}, which minimize $\|W - \widehat{W}\|$~\citep{Xu2018AlternatingMQ,Shang2023PB}, and 
(2) \emph{output-driven methods}, which align the outputs of the quantized model with those of the full-precision model by minimizing $\Vert \widehat{X} W - \widehat{X} \widehat{W} \Vert$~\citep{Li2024ARBLLMAR}, where $W$, $\widehat{W}$, and $\widehat{X}$ denote the full-precision weights, quantized weights, and quantized model's intermediate activations, respectively.
Despite the simplicity and stability, the weight-driven approach does not explicitly align with the primary objective of quantization, which is to preserve the full-precision model's final prediction. The output-driven approach is, therefore, more principled, as it minimizes the discrepancy in the model's outputs. However, naively applying output-driven objectives in the 1-bit regime leads to significant and often unexpected performance degradation. In this work, we identify that this failure arises from two key issues.

\noindent
\textbf{First, error accumulation across layers. }
In practice, existing binary LLM quantization frameworks \citep{Li2024ARBLLMAR} prioritize matching \(\widehat{X} \widehat{W}\) with the intermediate representations $\widehat{X} W$ rather than the original full-precision signals \(X W\) with \(X\) being the full-precision model's activation. However, as quantization errors accumulate, these local targets progressively drift from the true model behavior, creating a discrepancy between layer-wise optimization and the global end-to-end objective. \textbf{Although recent work \citep{Li2025GPTQv2EF} addresses error propagation during the error-compensation stage after each quantization decision, the same accumulated input mismatch also affects the quantization-variable optimization itself, which remains largely unexplored.} 


\noindent
\textbf{Second, magnitude bias and directional distortion in the optimization process.}
Simply minimizing mean squared error (MSE) in Euclidean space allocates the limited reconstruction capacity of a 1-bit model unevenly: dimensions with larger activation magnitude contribute more strongly to the objective and therefore receive disproportionately larger reconstruction improvements. More importantly, MSE measures only the magnitude of representation error and does not distinguish between perturbations that primarily change representation scale and those that change representation direction. This distinction is particularly important in pre-normalized Transformers, where subsequent computation repeatedly operates on normalized representations: normalization largely suppresses radial perturbations while preserving directional changes. Consequently, two quantization errors with similar MSE can have substantially different effects on the normalized representation. We find that unconstrained output alignment increasingly distorts the pairwise directional geometry among token representations, and that, even under matched reconstruction error, larger token-geometry distortion leads to larger downstream degradation.

Motivated by these observations, we propose a guided output-driven quantization method for 1-bit LLMs. Instead of relying on intermediate approximations, our method aligns quantized outputs with true full-precision targets, mitigating the impact of error accumulation. Furthermore, we formulate quantization as a geometry-constrained optimization problem that prevents reconstruction improvements from coming at the expense of token representation geometry. By constructing a local second-order approximation of the geometry constraint, we derive a closed-form projection that preserves token representation geometry while remaining close to the reconstruction-optimal update. These design choices yield a simple yet effective output-driven PTQ framework for 1-bit LLMs.  
\textbf{We focus on the binary setting, where the factorized binary parameterization admits efficient analytical conditional updates. Extending the proposed constrained formulation to multi-bit quantizers with discrete multi-level assignments is beyond the scope of this work.} In higher-bit quantization (e.g., 2-, 4-, or 8-bit), not all quantization variables admit analytic solutions; as a result, existing approaches in these settings typically rely on heuristic rounding schemes or gradient-based optimization
which falls outside the scope of this work. In summary, our contributions are as follows:
\begin{itemize}[topsep=0pt, itemsep=0pt, leftmargin=5ex]
\item We identify two fundamental limitations of output-driven quantization in the 1-bit regime: \emph{error accumulation} across layers and \emph{anisotropic distortion} of token representations under MSE-based reconstruction. Through controlled perturbations with matched reconstruction error, we further show that larger token-geometry distortion leads to larger downstream degradation.
\item We propose a novel PTQ method that optimizes true-target output reconstruction under a geometry-preserving constraint, mitigating the anisotropic distortion introduced by unconstrained output alignment. We further introduce a token-geometry-aware salient-mask selection strategy that identifies the most critical parameters for preserving token geometry.

\item Extensive experiments demonstrate that our method consistently outperforms existing 1-bit PTQ techniques for LLMs.
\end{itemize}

\section{Related works}

\noindent
\textbf{Quantization in LLMs.}
PTQ has emerged as a practical approach for compressing LLMs, as it can be applied directly to pretrained models with only a small calibration set, avoiding the high cost of Quantization-Aware Training (QAT). A range of PTQ methods have been proposed to improve the performance of quantized models further. For example,  SmoothQuant \citep{xiao2023smoothquant} mitigates activation outliers by applying scaling transformations that shift quantization difficulty from activations to weights. 
GPTQ \citep{frantar2022gptq} formulates quantization as a two-stage process consisting of a quantization stage followed by an error compensation stage, where second-order Hessian information is exploited for layer-wise error correction. Building on this framework, subsequent works such as GPTAQ \citep{Li2025GPTQv2EF} and QEP \citep{Arai2025QuantizationEP} aim to improve performance by addressing error propagation during the compensation stage. However, these methods primarily focus on mitigating error propagation after quantization, while paying relatively little attention to improving the initial quantization stage itself. More recent efforts such as QuIP\# \citep{tseng2024quip} and QuaRot \citep{ashkboos2024quarot} extend PTQ with rotation or vector quantization to better distribute outliers, though often at the expense of higher computational overhead. 
These efforts have helped LLMs maintain strong performance under moderate precision settings (e.g., 4-8 bits), yet the models still suffer from substantial degradation when pushed to extreme regimes. In this work, we focus on improving the binary quantization scheme through a closed-form formulation, without introducing additional inference computational overhead or auxiliary components.

\noindent
\textbf{1-Bit quantization for Large Language Models.}
Binarization, where weights are restricted to $\pm 1$, represents the most aggressive form of quantization. It was first explored in computer vision with specialized binary architectures such as XNOR-Net \citep{Rastegari2016XNORNetIC} and Bi-Real Net \citep{Liu2018BiRealNE}, which showed that binary parameters could still capture meaningful representations. Follow-up studies \citep{Guo2017NetworkSE,Xu2018AlternatingMQ} improved 1-bit quantization through enhanced coding schemes and optimized search strategies, enabling more accurate approximations of full-precision weights.  Inspired by these advances, recent work has extended binarization to LLMs. Training-based approaches, such as BitNet \citep{Wang2023BitNetS1}, demonstrated that end-to-end training with binary weights is feasible. In contrast, post-training quantization (PTQ) approaches aim to binarize pretrained models with minimal retraining. BiLLM \citep{huang2024billm} selectively quantizes salient weights with low-bit precision while binarizing the rest, guided by Hessian-based importance and residual-aware masks. STB-LLM \citep{Dong2024STBLLMBT} combines pruning and quantization with fine-grained grouping, achieving sub-1-bit average precision while maintaining accuracy, albeit with added kernel and storage costs. Most recently, research has shifted toward data-aware and fine-grained quantizers tailored for 1-bit PTQ. ARB \citep{Li2024ARBLLMAR} introduces grouping and refinement strategies to reduce quantization error, and its data-aware extension ARB-X further optimizes the output alignment.

\begin{figure*}[t]
    \centering

    \begin{subfigure}[t]{0.32\textwidth}
        \centering
        \includegraphics[width=\linewidth,height=3.6cm,keepaspectratio]
        {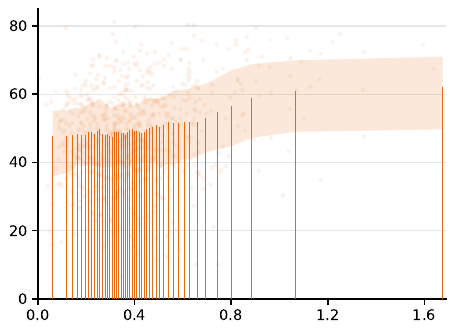}
        \caption{
        Uneven reconstruction improvement.
        Output-driven optimization preferentially improves high-magnitude
        representation channels.
        }
        \label{fig:error_propagation}
    \end{subfigure}
    \hfill
    \begin{subfigure}[t]{0.32\textwidth}
        \centering
        \includegraphics[width=\linewidth,height=3.6cm,keepaspectratio]
        {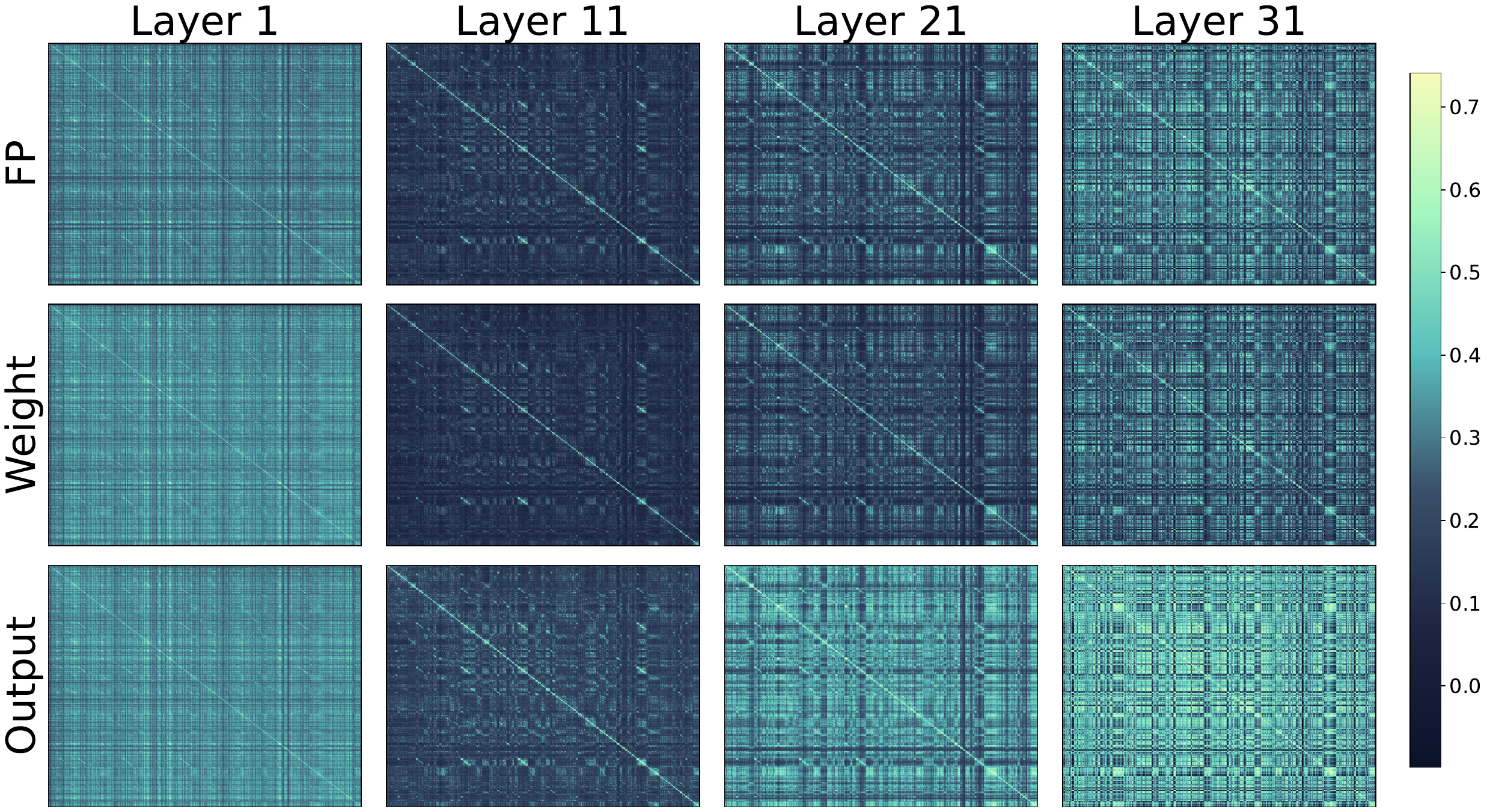}
        \caption{
        Accumulated geometry distortion.
        Output-driven optimization progressively deviates from the
        full-precision token geometry across layers.
        }
        \label{fig:token_similarity_drift}
    \end{subfigure}
    \hfill
    \begin{subfigure}[t]{0.32\textwidth}
        \centering
        \includegraphics[width=\linewidth,height=3.6cm,keepaspectratio]
        {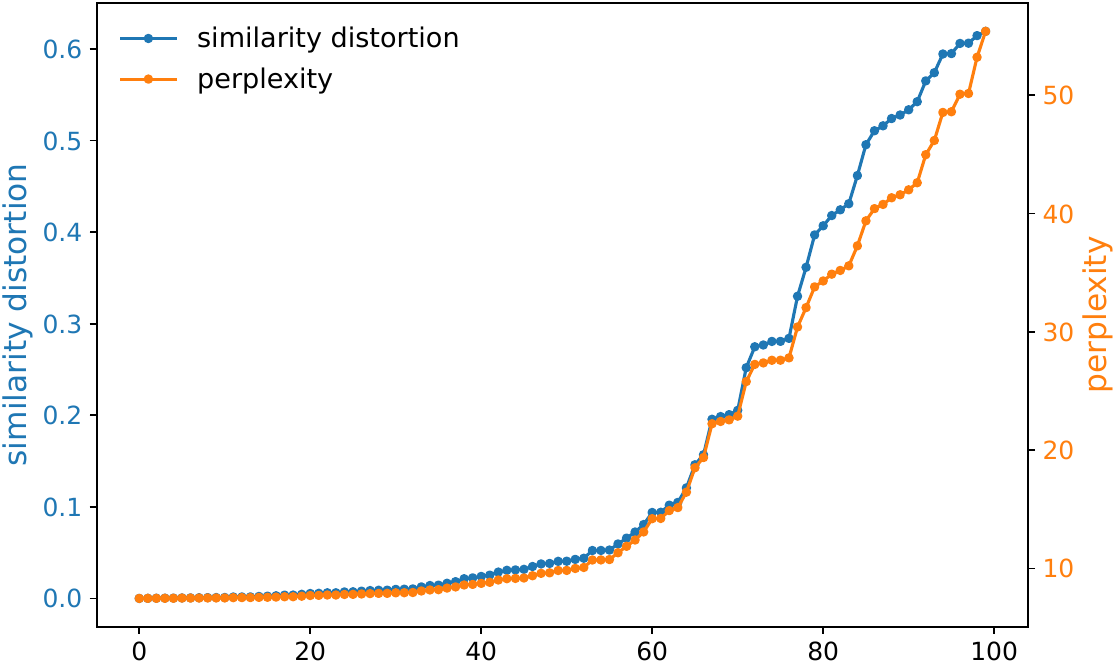}
        \caption{
        Geometry matters under controlled reconstruction error.
        At identical perturbation magnitude, larger token-geometry distortion
        leads to larger degradation.
        }
        \label{fig:controlled_geometry}
    \end{subfigure}

    \caption{
    \textbf{Reconstruction error alone does not characterize the functional
    distortion introduced by extreme quantization.}
    (a) Output-driven optimization allocates its reconstruction improvement
    unevenly across representation dimensions, disproportionately benefiting
    high-magnitude channels.
    (b) Despite reducing reconstruction error, it induces increasing distortion
    of the pairwise geometry between token representations.
    (c) We isolate this effect by perturbing the first-block representations
    with equal-magnitude errors while varying only their direction. Although
    all perturbations have identical reconstruction error, those that induce
    larger token-geometry distortion lead to higher final perplexity.
    Together, these observations suggest that preserving representation
    direction and token relationships captures information missed by
    reconstruction MSE.
    }
    \label{fig:combined_insights}
    \vspace{-0.1in}
\end{figure*}

\section{Preliminary Analysis}
\label{sec:preliminary}

\paragraph{Output reconstruction improves MSE unevenly and distorts token geometry.}
We first examine why output-driven quantization can achieve lower
reconstruction error yet inferior downstream performance in the extreme
1-bit regime. Given the full-precision output $Y^{\mathrm{FP}}$ and its
quantized counterpart $\widehat{Y}$, output-driven methods minimize the difference between $Y^{\mathrm{FP}}$ and $\widehat{Y}$:
\begin{equation}
    \mathcal{L}_{\mathrm{rec}}
    =
    \|Y^{\mathrm{FP}}-\widehat{Y}\|_F^2.
    \label{eq:reconstruction}
\end{equation}
Because this objective is dominated by dimensions with large absolute
errors, the limited reconstruction capacity of a 1-bit model need not be
allocated uniformly across the representation. We quantify this effect using the fractional error reduction of each output
channel $c$:
\begin{equation}
    R_c =
    \frac{
        E_c^{\mathrm{before}}-E_c^{\mathrm{after}}
    }{
        E_c^{\mathrm{before}}+\epsilon
    },
    \label{eq:relative_improvement}
\end{equation}
$E_{c} = \Vert Y^{\mathrm{FP}}_{:,c}-\widehat{Y}_{:,c} \Vert_{F}^{2}$ is the residual in channel $c$, and $\epsilon$ is a small positive number to avoid dividing by zero.
As shown in \cref{fig:error_propagation}, high-magnitude channels
receive substantially larger fractional error reductions, revealing a
strongly uneven allocation of reconstruction improvement.

Importantly, this reduction in MSE does not preserve the relational structure
between tokens. 
This can be observed via the token geometry using the normalized Gram matrix $G(Y)=\overline{Y} \, \overline{Y}^{\top}$ where \(\overline{Y}_{t, :} = \frac{Y_{t,:}}{\|Y_{t,:}\|_{F} + \epsilon}\) is the normalized output vector.
Each entry in the Gram matrix $G(Y)$ is the pairwise cosine similarity between token
representations. \cref{fig:token_similarity_drift} shows that
output-driven optimization progressively departs from the full-precision (FP)
token geometry, even as reconstruction error decreases, whereas
weight-driven quantization preserves this structure substantially better.
The discrepancy further increases with depth, suggesting that the
geometry distortion introduced by local reconstruction accumulates through
the network.

\paragraph{Why is reconstruction error insufficient?}
Reconstruction MSE measures the total magnitude of a perturbation but does
not distinguish between errors that change representation scale and those
that change representation direction. For a quantization error $\delta$,
we can decompose
\begin{equation}
    \delta = \delta_{\parallel} + \delta_{\perp},
    \qquad
    \|\delta\|_2^2
    =
    \|\delta_{\parallel}\|_2^2
    +
    \|\delta_{\perp}\|_2^2,
\end{equation}
where $\delta_{\parallel}$ is parallel to the original representation and
$\delta_{\perp}$ changes its direction. Reconstruction MSE penalizes both
components equally. This distinction is important in pre-normalized Transformers. Ignoring the
learned gain and numerical stabilizer, RMSNorm maps each token approximately
to $y/\|y\|_2$. It therefore largely suppresses scale-changing errors while
preserving errors that alter representation direction. Consequently, two
perturbations with identical reconstruction error can induce 
different normalized representations. The cosine geometry in $G(Y)$ directly captures these changes in relative token direction, providing information that  reconstruction alone cannot distinguish.

\paragraph{Does token geometry matter beyond reconstruction error?}
To isolate the effect of perturbation direction from its magnitude, we inject
$100$ independently sampled perturbations into the output of the first
Transformer block. Specifically, for each trial we sample
$ E_i$ with i.i.d.\ Gaussian entries and construct $\Delta_i
=
\eta
\frac{\|Y\|_F}{\|E_i\|_F}
 E_i,
\quad i=1,\ldots,100,$
such that $\frac{\|\Delta_i\|_F}{\| Y\|_F}
=
0.5 $ for all trials. We then continue the forward pass from
$ Y_i= Y+\Delta_i$ and measure both the resulting
token-geometry distortion and final perplexity.
Since all perturbations have identical Frobenius norm, their reconstruction
errors are exactly matched, while their directions differ.
As shown in Fig.~\ref{fig:controlled_geometry}, larger token-geometry distortion is
strongly associated with larger performance degradation even under matched
reconstruction error.

\section{Method}
\label{sec:method}

Our method uses token geometry as a common principle at two points in the
quantization pipeline. During parameter refinement, it filters
reconstruction-optimal updates through a geometry-preserving constraint.
Before refinement, it uses the same representation-level criterion to allocate
the limited residual binary capacity. We retain error-aware output
reconstruction as the optimization backbone and introduce the geometry terms
only where reconstruction error alone is insufficient. Consider a linear layer with full-precision weights
$W\in\mathbb R^{d_{\mathrm{in}}\times d_{\mathrm{out}}}$.
Let
$X,\widehat X\in\mathbb R^{N\times d_{\mathrm{in}}}$
denote the inputs generated by the full-precision and partially quantized
models, respectively. We parameterize the binary weight as
\begin{equation}
\label{eq:quantized_weight}
    \widehat W
    =
    \operatorname{Diag}(\alpha_r)
    \, B \,
    \operatorname{Diag}(\alpha_c),
\end{equation}
where
$B\in\{-1,+1\}^{d_{\mathrm{in}}\times d_{\mathrm{out}}}$,
$\alpha_r\in\mathbb R^{d_{\mathrm{in}}}$, and
$\alpha_c\in\mathbb R^{d_{\mathrm{out}}}$, and $\operatorname{Diag}(v)$ denotes a diagonal matrix constructed from vector $v$.

\subsection{Error-Aware Output Reconstruction}
\label{subsec:error_aware_reconstruction}

\paragraph{Pseudo-target reconstruction.}
Conventional layer-wise output alignment reconstructs the response of the
full-precision weight to the already perturbed activation:
\begin{equation}
\label{eq:pseudo_target_loss}
    \mathcal L_{\mathrm{pseudo}}(\widehat W)
    =
    \|\widehat XW-\widehat X\widehat W\|_F^2
    =
    \|\widehat X(W-\widehat W)\|_F^2.
\end{equation}
Although this objective provides a local regression target, it does not
explicitly account for the discrepancy between $X$ and $\widehat X$.
As quantization proceeds through the network, this discrepancy accumulates
and progressively shifts the layer-wise target away from the
full-precision target.

\paragraph{True-target reconstruction.}
We instead reconstruct the full-precision layer output:
\begin{equation}
\label{eq:reconstruction_loss}
    \mathcal L_Q(\widehat W)
    =
    \|XW-\widehat X\widehat W\|_F^2.
\end{equation}

The reconstruction loss in \cref{eq:reconstruction_loss} can then be rewritten in terms of the current weight approximation error and the inherited activation residual as follows:
\begin{equation}
\label{eq:expanded_reconstruction}
\begin{aligned}[b]
    L_{Q}(\widehat{W}) & = \Vert \widehat{X} (W - \widehat{W}) + (X - \widehat{X}) W \Vert_{F}^{2} \\
    & = \Vert \widehat{X} (W - \widehat{W}) \Vert_{F}^{2} + \Vert (X - \widehat{X}) W \Vert_{F}^{2} + 2 \langle \widehat{X} (W - \widehat{W}), (X - \widehat{X}) W \rangle_{F},
\end{aligned}
\end{equation}
where
$\langle U,V\rangle_F=\operatorname{Tr}(U^\top V)$. The first term in \cref{eq:expanded_reconstruction} measures the error introduced by the current quantized layer,
whereas the second term represents the error inherited from preceding layers
and is constant with respect to $\widehat W$. For compactness, we define \(\widehat{S} = \widehat{X}^{\top} \widehat{X}\) and \(S = \widehat{X}^{\top} X\) 
to rewrite the reconstruction loss as follows:
\begin{equation}
\label{eq:reconstruction_trace}
\begin{aligned}[b]
    \mathcal L_Q(\widehat W)
    ={}&
    \operatorname{Tr}(\widehat W^\top  \widehat{S} \widehat W)
    -
    2\operatorname{Tr}(\widehat W^\top SW) +
    \operatorname{Tr}(W^\top X^\top XW).
\end{aligned}
\end{equation}
Although \cref{eq:reconstruction_trace} is an unconstrained quadratic program w.r.t.
$\widehat W$, optimizing it under the coupled factorization in
\cref{eq:quantized_weight} is a mixed discrete-continuous problem.
We therefore optimize the quantization variables conditionally while holding
the remaining variables unchanged.

\subsection{Token-Geometry Regularization}
\label{subsec:geometry_regularization}

Our analysis in Section~\ref{sec:preliminary} characterizes token geometry through
the normalized pairwise similarities.
Directly optimizing this quantity introduces representation-dependent normalization,
breaking the blockwise quadratic structure required by our analytical binary updates.
We therefore use the following cross-Gram affinity as a tractable surrogate:


\begin{equation}
\label{eq:geometry_regularizer}
\begin{aligned}[b]
    \hspace*{-1ex}
    \mathcal S_{\mathrm{Reg}}(\widehat W)
    &=
    \nicefrac{1}{2}
    \langle
        (\widehat X\widehat W)
        (\widehat X\widehat W)^\top,
        (XW)(XW)^\top
    \rangle_F
    =
    \nicefrac{1}{2}
    \operatorname{Tr}
    \left(
        \widehat W^\top M\widehat W
    \right),
\end{aligned}
\end{equation}
where $M = \widehat X^\top X W W^\top X^\top\widehat X = S W W^\top S^\top$. We use this cross-Gram affinity as a tractable surrogate for token geometry: it captures agreement between all pairs of token outputs while remaining quadratic in each block of quantization variables. In the Appendix \ref{sec:true_geometry_target}, we provide visualizations showing that optimizing this surrogate consistently improves the target geometry measures. A larger $\mathcal S_{\mathrm{Reg}}$ indicates stronger affinity between the
pairwise inner-product structures of the full-precision and quantized token
representations.
Accordingly, we optimize the reconstruction objective while requiring the
geometry-alignment score not to decrease:
\begin{equation}
\label{eq:geometry_constrained_objective}
\begin{aligned}
    \textstyle \min_{\alpha_r,\alpha_c,B}\,\mathcal L_Q(\widehat W) \quad \text{s.t.: } \mathcal S_{\mathrm{Reg}}(\widehat W)
        \geq
        \mathcal S_{\mathrm{Reg}}(\widehat W^{(t)}),
\end{aligned}
\end{equation}
where $\widehat W^{(t)}$ denotes the current quantized weights. The
reconstruction term controls output scale, while the constraint prevents an
accepted update from improving reconstruction by sacrificing relational
structure.
Thus, the reconstruction objective determines the preferred update, while
the geometry constraint restricts updates that distort token relationships.

The joint problem contains continuous scales, discrete binary variables, and
a coupled quadratic constraint, hence it has no joint closed-form solution. We optimize one variable block $\phi
    \in
    \left\{
        \alpha_c,\alpha_r,B_{i,:}
    \right\}$ at a time, while keeping all remaining variables fixed.

\subsection{Reconstruction-Optimal Proposals}
\label{subsec:reconstruction_proposals}

For a current variable block $\phi$, we first compute its
reconstruction closed-form optimum for $\mathcal{L}_{Q}$ as 
\(\phi_Q^\star \in \operatorname*{\arg\min}_{\phi}\mathcal L_Q(\phi)\) (see \cref{sec:derivation_closedform}).
The update $\Delta_\phi^Q 
= \phi_Q^\star - \phi $ is the analytical closed-form solution derived by minimizing $\mathcal{L}_{Q}$ while keeping other variables fixed. This update step may violate the constraint in
\cref{eq:geometry_constrained_objective}. We therefore seek a feasible update that
remains close to the reconstruction optimum. To retain the analytical nature
of the proposal, each coordinate either takes its complete proposed change or
remains unchanged:
\begin{equation}
\label{eq:discrete_projection_original}
\begin{aligned}[b]
    \textstyle
    \min_{\Delta}\quad
        \nicefrac{1}{2}\Vert \Delta-\Delta_\phi^Q \Vert^2,
    \quad\text{s.t.: } \begin{cases}
        \mathcal{S}_{\mathrm{Reg}}(\phi+\Delta)
        \geq \mathcal{S}_{\mathrm{Reg}}(\phi),\\
        \Delta_j \in \{0,(\Delta_\phi^Q)_j \},
        \forall j.
        \end{cases}
\end{aligned}
\end{equation}

\subsection{Local Second-Order Geometry Constraint}
\label{subsec:second_order_geometry}

The exact quadratic geometry constraint in
\cref{eq:discrete_projection_original} couples the coordinates of $\Delta$,
making the discrete projection intractable. We obtain a separable local
surrogate by expanding $\mathcal{S}_{\mathrm{Reg}}$ around the current
variable block $\phi$ and retaining only diagonal curvature:
\begin{equation}
\label{eq:geometry_taylor}
\begin{aligned}[b]
    &\mathcal{S}_{\mathrm{Reg}}(\phi+\Delta) - \mathcal{S}_{\mathrm{Reg}}(\phi)
    =
    g_\phi^\top\Delta
    +
    \nicefrac{1}{2}
    \Delta^\top H_\phi\Delta
    \textstyle \approx
    \sum_j
    \left[
        (g_\phi)_j\Delta_j
        +
        \nicefrac{\beta}{2}
        (h_\phi)_j\Delta_j^2
    \right],
\end{aligned}
\end{equation}
where $g_\phi
    =
    \nabla_\phi\mathcal{S}_{\mathrm{Reg}}(\phi)$ is the gradient, $
    H_\phi
    =
    \nabla_\phi^2\mathcal{S}_{\mathrm{Reg}}(\phi)$ is the Hessian matrix,
    $h_\phi
    =
    \operatorname{diag}(H_\phi)$ is the Hessian diagonal. The coefficient
$\beta\geq0$ controls the strength of this term. To obtain a separable feasible region, we impose the local constraint in \cref{eq:geometry_taylor} on every coordinate:
\begin{equation}
\label{eq:coordinate_geometry_constraint}
    (g_\phi)_j\Delta_j
    +
    \nicefrac{\beta}{2}
    (h_\phi)_j\Delta_j^2
    \geq 0,
    \forall j.
\end{equation}
Substituting this approximation into
\cref{eq:geometry_constrained_objective} gives:
\begin{equation}
\label{eq:separable_projection_problem}
\begin{aligned}[b]
    \textstyle \min_{\Delta}\,
        \nicefrac{1}{2}
        \|\Delta-\Delta_\phi^Q\|^2 \quad 
    \text{s.t.: } \begin{cases}
        (g_\phi)_j\Delta_j
        +
        \nicefrac{\beta}{2}
        (h_\phi)_j\Delta_j^2
        \geq 0,
        \forall j,\\
        \Delta_j
        \in
        \left\{0,(\Delta_\phi^Q)_j\right\},
        \forall j.
        \end{cases}
\end{aligned}
\end{equation}

\subsection{Closed-Form Geometry-Preserving Projection}
\label{subsec:closed_form_projection}

For each coordinate $j$ in \cref{eq:separable_projection_problem}, retaining $(\Delta_\phi^Q)_j$ produces zero projection
error and is optimal whenever it satisfies the local geometry constraint.
Otherwise, the only remaining feasible choice is zero. We therefore define
\begin{equation}
\label{eq:geometry_mask}
    (M_\phi)_j
    =
    \mathbf 1
    \left[
        (g_\phi)_j(\Delta_\phi^Q)_j
        +
        \nicefrac{\beta}{2}
        (h_\phi)_j(\Delta_\phi^Q)_j^2
        \geq0
    \right],
\end{equation}
where $\mathbf{1}(.) \in \{0, 1\}$ is the indicator function.
The projected update is
\begin{equation}
\label{eq:projected_update}
    \widetilde\Delta_\phi
    =
    M_\phi\odot\Delta_\phi^Q,
    \qquad
    \widetilde\phi
    =
    \phi+\widetilde\Delta_\phi.
\end{equation}
\cref{prop:geometry_preservation} states the local property for constructing
the candidate update (see \cref{app:amp_proof} for the proof).
\begin{proposition}[Local geometry preservation]
\label{prop:geometry_preservation}
Under the diagonal-curvature approximation in
\cref{eq:geometry_taylor}, the projected update does not decrease the
local geometry-alignment model:
\[
\textstyle \sum_j\left[(g_\phi)_j\widetilde\Delta_{\phi,j}
+\frac{\beta}{2}(h_\phi)_j\widetilde\Delta_{\phi,j}^2\right]\geq0.
\]
\end{proposition}

The update in \cref{eq:projected_update} is the exact solution of the
\emph{discrete separable local projection} in
\cref{eq:separable_projection_problem}. Because diagonalization is an
approximation, we accept the candidate only when the actual reconstruction
loss does not increase and the actual score
$\mathcal S_{\mathrm{Reg}}$ does not decrease. This final check enforces the
original constraint in \cref{eq:discrete_projection_original}.

\subsection{Conditional Reconstruction Solutions}
\label{subsec:conditional_solutions}

Our optimization alternates over each quantization variable
$\phi \in \{\alpha_r,\alpha_c,B\}$ while keeping the remaining variables
fixed. Each subproblem admits an
analytical closed-form update, which is used to compute
$\Delta_\phi^Q$. We also provide the geometry terms, $g_{\phi}$ and $h_{\phi}$, used by the
proposed constrained optimization. The complete derivations are provided in
\cref{app:analytical_solution,sec:derivation_closedform}.
\subsection{Token-Geometry-Guided Mask Selection}
\label{subsec:geometry_mask_search}

For each layer, the binary quantizer uses two binary matrices to represent a
subset of salient columns and one binary matrix for the remaining columns.
Rather than determining this subset solely from weight reconstruction error, we
use token geometry to determine the final allocation. Following the saliency
criterion of the underlying binary quantizer, we define the 
full-precision activation Gram matrix
$H = X^{\top}X + \epsilon\mathbf I$
and let $U$ denote the upper-triangular Cholesky factor satisfying
$H^{-1} = U^{\top}U$.
We rank the $d_{\mathrm{in}}$ input columns according to
\begin{equation}
s_j
=
\nicefrac{
\sum_i
\left(
W_{ij}
\right)^2
}{
U_{jj}^2
},
\label{eq:saliency_score}
\end{equation}
and denote the resulting descending order by
$\{c_1,\ldots,c_{d_{\mathrm{in}}}\}$.
For each quantization group and each prefix
$\mathcal{C}_m = \{c_1,\ldots,c_m\}$,
$1\leq m<K$, let $\widetilde{W}^{(m)}_q$ denote the pseudo-quantized weight
obtained when the columns in $\mathcal{C}_m$ receive the salient two-term
representation. Among these $K-1$ candidates, we select the mask whose induced
token geometry is closest to that of the full-precision output:
\begin{equation}
\textstyle
m^\star
=
\arg\min_{1\leq m<K}
\frac{
\left\|
Z_q^{(m)} Z_q^{(m)\top}
-
ZZ^\top
\right\|_F^2
}{
\left\|
ZZ^\top
\right\|_F^2
+\epsilon
}.
\label{eq:token_geometry_mask_search}
\end{equation}
Here, $Z$ and $Z_q^{(m)}$ are respectively the FP output and the
pseudo-quantized output after projection to a lower-dimensional space.
Specifically, we use $P\in\mathbb{R}^{d_{\mathrm{out}}\times r}$ with orthonormal columns, where $r\ll d_{\mathrm{out}}$, and construct $P$ by sampling a Gaussian random matrix and orthonormalizing its columns using reduced QR decomposition.  The same projection is used for all candidate masks within the group. The selected prefix $\mathcal{C}_{m^\star}$ defines the salient columns. Since $m^\star$ is selected independently for each group, $K$ bounds the candidate range and is chosen such that the resulting average bit-width remains comparable to the underlying baseline. This converts mask selection into an explicit token-geometry preservation problem.

\subsection{Overall Algorithm}
\label{subsec:overall_algorithm}
We initialize the binary weights with the signs of the full-precision weights,
as in conventional binary quantization. We apply output-driven refinement to the final fully connected layer of each Transformer block, with the remaining linear layers following ARB-RC. We then select the two-term residual
mask using \cref{subsec:geometry_mask_search} and iteratively refine the
quantization parameters using the geometry-preserving projection. Thus, token
geometry governs both capacity allocation and update acceptance. The complete procedure
for iterative refinement is provided in \cref{app:learning_algorithm}.

\section{Experiments}

\begin{table*}[t]
\centering
\caption{
Perplexity $(\downarrow)$ results of OPT and LLaMA models on C4,
WikiText2, and PTB. Separate weight-bit columns are used for each
model family. \textbf{Use OA} denotes methods that use an
output-alignment objective for quantization.
}
\label{tab:combined_opt_llama}
{\small
\setlength{\tabcolsep}{4.6pt}
\renewcommand{\arraystretch}{1.02}

\resizebox{\linewidth}{!}{
\begin{tabular}{c|c|c|c|ccc|c|ccc}
\toprule
\multirow{2}{*}{\textbf{Dataset}}
& \multirow{2}{*}{\textbf{Method}}

& \textbf{Use}

& \textbf{OPT}
& \multicolumn{3}{c|}{\textbf{OPT}}
& \textbf{LLaMA}
& \multicolumn{3}{c}{\textbf{LLaMA}} \\
\cmidrule(lr){5-7}
\cmidrule(lr){9-11}
& 
& \textbf{OA}
& \textbf{Bits}
& \textbf{1.3B}
& \textbf{2.7B}
& \textbf{6.7B}
& \textbf{Bits}
& \textbf{LLaMA-2 7B}
& \textbf{LLaMA-2 13B}
& \textbf{LLaMA-3 8B} \\
\midrule


\rowcolor{lightgray}
\multirow{8}{*}{\rotatebox{90}{\textbf{C4}}}
& Full Precision & --
& 16
& 16.07 & 14.34 & 12.71
& 16
& 7.26 & 6.73 & 9.45 \\

& PB-LLM & \(\textcolor{Red}{\times}\)
& 1.7
& 168.12 & 222.15 & 104.78
& 1.7
& 80.69 & 184.67 & 104.15 \\

& BiLLM & \(\textcolor{Red}{\times}\)
& 1.11
& 64.14 & 44.77 & 42.13
& 1.08
& 39.38 & 25.87 & 61.04 \\

& ARB-X & \(\textcolor{PineGreen}{\checkmark}\)
& 1.11
& 47.60 & 34.97 & 22.54
& 1.08
& 28.02 & 19.82 & 41.86 \\

& OAC & \(\textcolor{Red}{\times}\)
& 1.11
& -- & -- & --
& 1.08
& 21.64 & 15.25 & -- \\

& DBellQuant & \(\textcolor{Red}{\times}\)
& 1.11
& 42.86 & 33.19 & 21.75
& 1.08
& 21.83 & 15.14 & -- \\

& ARB-RC & \(\textcolor{Red}{\times}\)
& 1.11
& $28.03 \pm 0.30$
& $21.51 \pm 0.15$
& $17.02 \pm 0.10$
& 1.08
& $20.56 \pm 0.68$
& $14.76 \pm 0.50$
& $36.04 \pm 0.01$ \\

\rowcolor{midgray}
& \textbf{Ours} & \(\textcolor{PineGreen}{\checkmark}\)
& 1.11
& $\mathbf{24.65 \pm 0.05}$
& $\mathbf{19.90 \pm 0.06}$
& $\mathbf{16.18 \pm 0.03}$
& 1.08
& $\mathbf{17.74 \pm 0.30}$
& $\mathbf{13.69 \pm 0.11}$
& $\mathbf{35.16 \pm 0.20}$ \\

\midrule


\rowcolor{lightgray}
\multirow{8}{*}{\rotatebox{90}{\textbf{WikiText2}}}
& Full Precision & --
& 16
& 14.62 & 12.47 & 10.86
& 16
& 5.47 & 4.88 & 6.14 \\

& PB-LLM & \(\textcolor{Red}{\times}\)
& 1.7
& 239.81 & 278.27 & 144.25
& 1.7
& 66.41 & 236.40 & 73.08 \\

& BiLLM & \(\textcolor{Red}{\times}\)
& 1.11
& 69.05 & 48.61 & 47.65
& 1.08
& 32.31 & 21.35 & 55.80 \\

& ARB-X & \(\textcolor{PineGreen}{\checkmark}\)
& 1.11
& 45.40 & 34.37 & 20.07
& 1.08
& 21.61 & 14.86 & 31.98 \\

& OAC & \(\textcolor{Red}{\times}\)
& 1.11
& -- & -- & --
& 1.08
& 19.50 & 13.15 & -- \\

& DBellQuant & \(\textcolor{Red}{\times}\)
& 1.11
& 43.42 & 31.47 & 18.89
& 1.08
& 17.91 & 12.79 & -- \\

& ARB-RC & \(\textcolor{Red}{\times}\)
& 1.11
& $26.83 \pm 0.15$
& $20.09 \pm 0.19$
& $15.06 \pm 0.08$
& 1.08
& $16.43 \pm 0.06$
& $12.33 \pm 0.15$
& $28.14 \pm 0.55$ \\

\rowcolor{midgray}
& \textbf{Ours} & \(\textcolor{PineGreen}{\checkmark}\)
& 1.11
& $\mathbf{24.11 \pm 0.25}$
& $\mathbf{18.22 \pm 0.03}$
& $\mathbf{14.28 \pm 0.18}$
& 1.08
& $\mathbf{14.62 \pm 0.05}$
& $\mathbf{10.98 \pm 0.19}$
& $\mathbf{26.62 \pm 0.29}$ \\

\midrule


\rowcolor{lightgray}
\multirow{6}{*}{\rotatebox{90}{\textbf{PTB}}}
& Full Precision & --
& 16
& 20.29 & 17.97 & 15.77
& 16
& 37.91 & 50.93 & 11.18 \\

& PB-LLM & \(\textcolor{Red}{\times}\)
& 1.7
& 324.62 & 183.97 & 169.49
& 1.7
& 657.24 & 816.31 & 106.25 \\

& BiLLM & \(\textcolor{Red}{\times}\)
& 1.11
& 115.94 & 88.52 & 69.41
& 1.08
& 5243.01 & 309.12 & 87.25 \\

& ARB-X & \(\textcolor{PineGreen}{\checkmark}\)
& 1.11
& 71.96 & 54.28 & 31.23
& 1.08
& 783.66 & 208.06 & 62.69 \\

& ARB-RC & \(\textcolor{Red}{\times}\)
& 1.11
& $42.91 \pm 0.20$
& $31.20 \pm 0.07$
& $22.21 \pm 0.11$
& 1.08
& $853.30 \pm 196.57$
& $196.31 \pm 2.05$
& $47.86 \pm 1.47$ \\

\rowcolor{midgray}
& \textbf{Ours} & \(\textcolor{PineGreen}{\checkmark}\)
& 1.11
& $\mathbf{37.03 \pm 0.40}$
& $\mathbf{28.69 \pm 0.51}$
& $\mathbf{21.24 \pm 0.42}$
& 1.08
& $\mathbf{248.56 \pm 45.39}$
& \textbf{188.54} \(\pm\) \textbf{1.6}
& \textbf{40.16} \(\pm\) \textbf{1.3 }\\

\bottomrule
\end{tabular}
}
}
\end{table*}

\noindent
\textbf{Models and datasets.}
Our experiments are conducted on the OPT model family~\citep{zhang2022opt} as well as on the LLaMA family, covering parameter scales from 1.3B to 13B, including LLaMA-2~\citep{touvron2023llama} and LLaMA-3~\citep{grattafiori2024llama}.
We also report results on newer architectures from the Qwen family. For evaluation, \emph{perplexity} is reported on WikiText2 \citep{merity2016pointer}, C4 \citep{raffel2020exploring} and PTB \citep{marcus-etal-1993-building}, which are standard for measuring language modeling quality. To assess downstream capability, we also measure zero-shot performance on eight QA reasoning datasets:  ARC-Easy and ARC-Challenge~\citep{arc}, PIQA~\citep{bisk2020piqa}, 
HellaSwag~\citep{hellaswag}, 
WinoGrande~\citep{sakaguchi2021winogrande} and OBQA \citep{OBQA}, COPA\citep{copa} and LAMBADA \citep{paperno2016lambada}. 

\noindent
\textbf{Experimental setting.} All experiments are implemented in PyTorch and executed on a single NVIDIA A100 GPU. To be consistent with prior studies such as GPTQ~\citep{frantar2022gptq} and BiLLM~\citep{huang2024billm}, we use 128 samples from the C4 dataset with a sequence length of 2048 as calibration data to enable fair comparison. The quantization block size is fixed at 128, and the optimization of quantization variables is performed with 15 iterations to be consistent with ARB \citep{Li2024ARBLLMAR}, $\beta$ is set to 1 for simplicity. Following the baselines, we only quantize the model weights, keeping the activation at full-precision. For token-geometry-aware mask selection, we set $K=25$ for Llama and Qwen while set $K=30$ for OPT, and set $r=32$ for all experiments.

\noindent
\textbf{Baseline methods.}
We compare our method against several state-of-the-art 1-bit PTQ methods, including
BiLLM \citep{huang2024billm}, ARB-LLM\citep{Li2024ARBLLMAR} and PB-LLM~\citep{Shang2023PB}, ensuring that all implementations adhere to the details provided in
their respective papers. BiLLM \citep{huang2024billm}, ARB-LLM\citep{Li2024ARBLLMAR}, PB-LLM \citep{Shang2023PB} and DBellQuant \citep{Ye2025DBellQuantBT} all utilize the PTQ approach for model
 calibration of GPTQ. For controlled comparison, we primarily consider 1-bit PTQ methods that retain the standard binary-weight representation without introducing additional representation transforms.  
 ARB-RC results and our results in Tables \ref{tab:combined_opt_llama} were obtained by running with 3 different random seeds.

\subsection{Experimental results}

\begin{table*}[t]
  \centering
  \caption{
  Zero-shot accuracy (\%) on QA reasoning benchmarks.
  Green values denote the improvement of our method
  over ARB-RC in percentage points.
  $^{\dagger}$The average is computed over all tasks.
  }
  \label{tab:llama_reasoning_results}
  \setlength{\tabcolsep}{4.0pt}
  \renewcommand{\arraystretch}{1.16}

  \resizebox{\textwidth}{!}{%
  \begin{tabular}{
      @{} c l c
      *{8}{c}
      >{\columncolor{avggray}}c
      @{}
  }
  \toprule
  \textbf{Model}
  & \textbf{Method}
  & \textbf{Bits}
  & \textbf{Lambada}
  & \textbf{PIQA}
  & \textbf{OBQA}
  & \textbf{WinoGrande}
  & \textbf{ARC-E}
  & \textbf{ARC-C}
  & \textbf{HellaSwag}
  & \textbf{COPA}
  & \textbf{Avg.}$^{\dagger}$ \\
  \midrule

  \multirow{3}{*}{\textbf{LLaMA-2-7B}}
    & ARB-RC
    & 1.08
    & 51.66
    & 64.85
    & 19.40
    & 59.51
    & 50.97
    & 24.32
    & 36.79
    & 71.00
    & 47.31 \\

    & DBELL
    & 1.08
    & 41.63
    & 64.85
    & 17.60
    & 56.91
    & 50.38
    & 23.98
    & 35.65
    & 63.00
    & 44.25 \\

    & \ours
    & 1.08
    & \textbf{58.53}\gain{6.87}
    & \textbf{67.46}\gain{2.61}
    & \textbf{20.80}\gain{1.40}
    & \textbf{60.14}\gain{0.63}
    & \textbf{53.41}\gain{2.44}
    & \textbf{25.60}\gain{1.28}
    & \textbf{38.91}\gain{2.12}
    & \textbf{74.00}\gain{3.00}
    & \cellcolor{gainbg}\textbf{49.86}\gain{2.55} \\
    \midrule

    \multirow{3}{*}{\textbf{LLaMA-2-13B}}
    & ARB-RC
    & 1.08
    & 66.02
    & 69.59
    & 23.20
    & 61.48
    & 61.28
    & 29.18
    & 38.73
    & 81.00
    & 53.81 \\

    & DBELL
    & 1.08
    & 55.75
    & 68.44
    & 20.80
    & 61.64
    & 60.10
    & 28.16
    & 39.08
    & 78.00
    & 51.50 \\

    & \ours
    & 1.08
    & \textbf{69.28}\gain{3.26}
    & \textbf{71.71}\gain{2.12}
    & \textbf{25.00}\gain{1.80}
    & \textbf{66.22}\gain{4.74}
    & \textbf{64.44}\gain{3.16}
    & \textbf{30.89}\gain{1.71}
    & \textbf{43.09}\gain{4.36}
    & \textbf{82.00}\gain{1.00}
    & \cellcolor{gainbg}\textbf{56.58}\gain{2.77} \\
    \midrule

    \multirow{3}{*}{\textbf{LLaMA-3-8B}}
    & ARB-RC
    & 1.08
    & 48.85
    & 61.43
    & 20.20
    & 57.46
    & 45.83
    & 22.01
    & 35.39
    & 76.00
    & 45.90 \\

    & DBELL
    & 1.08
    & 34.78
    & 59.30
    & 17.80
    & 54.85
    & 36.78
    & 18.86
    & 33.40
    & 73.00
    & 41.10 \\

    & \ours
    & 1.08
    & \textbf{57.73}\gain{8.88}
    & \textbf{67.19}\gain{5.76}
    & \textbf{24.60}\gain{4.40}
    & \textbf{59.35}\gain{1.89}
    & \textbf{56.82}\gain{10.99}
    & \textbf{26.62}\gain{4.61}
    & \textbf{38.40}\gain{3.01}
    & \textbf{77.00}\gain{1.00}
    & \cellcolor{gainbg}\textbf{50.96}\gain{5.06} \\

  \bottomrule
  \end{tabular}%
  }
  \end{table*}

\noindent
\textbf{Results on OPT and LLaMA models.}
As shown in Table~\ref{tab:combined_opt_llama}, our method consistently outperforms ARB-RC across all 18 perplexity comparisons on OPT and LLaMA models. The improvement is particularly pronounced in challenging settings, reducing PTB perplexity from 853.30 to 248.56 on LLaMA-2-7B and from 47.86 to 40.16 on LLaMA-3-8B. For downstream reasoning tasks, our method improves the average zero-shot accuracy over ARB-RC by 2.55, 2.77, and 5.06 \% points on LLaMA-2-7B, LLaMA-2-13B, and LLaMA-3-8B, respectively, and outperforms ARB-RC on all 24 individual benchmarks. The largest gain reaches 10.99 \% in accuracy on ARC-E for LLaMA-3-8B, demonstrating the effectiveness of the method.
 
\noindent
\textbf{Performance of our method over Qwen models.}
\label{subsec:qwen}
We further evaluate Qwen2, Qwen2.5, and Qwen3 to examine whether the benefit of geometry preservation transfers across model architectures. As shown in Table~\ref{tab:qwen2_2p5_combined}, our method improves over ARB-RC in all 15 perplexity comparisons and all five average QA comparisons. The gains are substantial on Qwen2.5-1.5B, where perplexity decreases from 128.77 to 64.33 on C4, from 107.92 to 48.11 on WikiText2, and from 175.40 to 96.48 on PTB, while average QA accuracy improves by 3.79 percentage points. These results show that preserving token geometry provides consistent benefits beyond a single model family.

\begin{table*}[!t]
\caption{
    Perplexity ($\downarrow$) on C4, WikiText2, and PTB, and average
    zero-shot reasoning accuracy ($\uparrow$) for Qwen models.
    All methods use a block size of 128.
    Green annotations report the absolute improvement over ARB-RC.
    }
    \label{tab:qwen2_2p5_combined}
    \centering
    \setlength{\tabcolsep}{4.5pt}
    \renewcommand{\arraystretch}{1.15}

    \resizebox{\linewidth}{!}{%
    \begin{tabular}{c|c|c|c|cc|cc|c}
    \toprule
    \multirow{2}{*}{\textbf{Dataset}} &
    \multirow{2}{*}{\textbf{Metric}} &
    \multirow{2}{*}{\textbf{Method}} &
    \multirow{2}{*}{\textbf{Block Size}} &
    \multicolumn{2}{c|}{\textbf{Qwen2}} &
    \multicolumn{2}{c|}{\textbf{Qwen2.5}} &
    \textbf{Qwen3} \\

    \cmidrule(lr){5-6}
    \cmidrule(lr){7-8}
    \cmidrule(lr){9-9}

    & & & &
    \textbf{1.5B} &
    \textbf{7B} &
    \textbf{1.5B} &
    \textbf{7B} &
    \textbf{8B} \\
    \midrule

    \multirow{3}{*}{\textbf{C4}}
    & \multirow{3}{*}{PPL ($\downarrow$)}
    & ARB-RC
    & 128
    & 109.83
    & 22.63
    & 128.77
    & 24.32
    & 32.71 \\

    &
      & DBellQuant
      & 128
      & 607.06
      & 29.03
      & 1428.88
      & 29.52
      & 50.88 \\

    &
    & \cellcolor{midgray}\textbf{Ours}
    & \cellcolor{midgray}128
    & \cellcolor{midgray}
      \textbf{74.83}\,
      {\scriptsize\color{gaincolor}($\downarrow$35.00)}
    & \cellcolor{midgray}
      \textbf{20.54}\,
      {\scriptsize\color{gaincolor}($\downarrow$2.09)}
    & \cellcolor{midgray}
      \textbf{64.33}\,
      {\scriptsize\color{gaincolor}($\downarrow$64.44)}
    & \cellcolor{midgray}
      \textbf{20.95}\,
      {\scriptsize\color{gaincolor}($\downarrow$3.37)}
    & \cellcolor{midgray}
      \textbf{27.68}\,
      {\scriptsize\color{gaincolor}($\downarrow$5.03)} \\
    \midrule

    \multirow{3}{*}{\textbf{WikiText2}}
    & \multirow{3}{*}{PPL ($\downarrow$)}
    & ARB-RC
    & 128
    & 89.59
    & 16.99
    & 107.92
    & 17.08
    & 26.03 \\


      &
      & DBellQuant
      & 128
      & 542.06
      & 28.52
      & 2416.90
      & 25.10
      & 58.16 \\

    &
    & \cellcolor{midgray}\textbf{Ours}
    & \cellcolor{midgray}128
    & \cellcolor{midgray}
      \textbf{68.45}\,
      {\scriptsize\color{gaincolor}($\downarrow$21.14)}
    & \cellcolor{midgray}
      \textbf{14.55}\,
      {\scriptsize\color{gaincolor}($\downarrow$2.44)}
    & \cellcolor{midgray}
      \textbf{48.11}\,
      {\scriptsize\color{gaincolor}($\downarrow$59.81)}
    & \cellcolor{midgray}
      \textbf{14.42}\,
      {\scriptsize\color{gaincolor}($\downarrow$2.66)}
    & \cellcolor{midgray}
      \textbf{20.73}\,
      {\scriptsize\color{gaincolor}($\downarrow$5.30)} \\
    \midrule

    \multirow{3}{*}{\textbf{PTB}}
    & \multirow{3}{*}{PPL ($\downarrow$)}
    & ARB-RC
    & 128
    & 135.49
    & 28.33
    & 175.40
    & 29.47
    & 44.56 \\

    &
      & DBellQuant
      & 128
      & 1246.88
      & 47.27
      & 3843.26
      & 43.99
      & 86.66 \\

    &
    & \cellcolor{midgray}\textbf{Ours}
    & \cellcolor{midgray}128
    & \cellcolor{midgray}
      \textbf{107.04}\,
      {\scriptsize\color{gaincolor}($\downarrow$28.45)}
    & \cellcolor{midgray}
      \textbf{23.50}\,
      {\scriptsize\color{gaincolor}($\downarrow$4.83)}
    & \cellcolor{midgray}
      \textbf{96.48}\,
      {\scriptsize\color{gaincolor}($\downarrow$78.92)}
    & \cellcolor{midgray}
      \textbf{25.24}\,
      {\scriptsize\color{gaincolor}($\downarrow$4.23)}
    & \cellcolor{midgray}
      \textbf{33.22}\,
      {\scriptsize\color{gaincolor}($\downarrow$11.34)} \\
    \midrule

      \multirow{3}{*}{\textbf{AveQA}}
      & \multirow{3}{*}{Acc. ($\uparrow$)}
      & ARB-RC
      & 128
      & 35.62
      & 53.95
      & 37.83
      & 54.79
      & 53.56 \\

      &
      & DBellQuant
      & 128
      & 31.53
      & 47.01
      & 32.17
      & 49.02
      & 43.84 \\

      &
      & \cellcolor{midgray}\textbf{Ours}
      & \cellcolor{midgray}128
      & \cellcolor{midgray}\textbf{38.32}\gain{2.70}
      & \cellcolor{midgray}\textbf{56.03}\gain{2.08}
      & \cellcolor{midgray}\textbf{41.62}\gain{3.79}
      & \cellcolor{midgray}\textbf{55.89}\gain{1.10}
      & \cellcolor{midgray}\textbf{55.22}\gain{1.65} \\

    \bottomrule
    \end{tabular}%
    }

  \end{table*}

\subsection{Ablation study}

\label{subsec:acivation_acucmulation}

\noindent
\textbf{Ablation of the constrained optimization objective.}

\begin{wraptable}{r}{0.5\linewidth}
\vspace{-2ex}
\caption{Ablation study about the contribution of each component, using a fixed random seed.}
\label{tab:true_target_geometry_ablation}
\centering

{\small
\setlength{\tabcolsep}{3.5pt}
\renewcommand{\arraystretch}{1.10}

\resizebox{\linewidth}{!}{

\begin{tabular}{ccc cc cc}
\toprule
\multicolumn{3}{c}{\textbf{Optimization}}
&
\multicolumn{2}{c}{\textbf{Qwen-2.5-1.5B}}
&
\multicolumn{2}{c}{\textbf{LLaMA-2-7B}}
\\
\cmidrule(lr){1-3}
\cmidrule(lr){4-5}
\cmidrule(lr){6-7}

$\mathcal{L}_{Q}$
&
$\Delta\mathcal{S}_{\mathrm{Reg}}\geq 0$
&
\textbf{Mask}
&
\textbf{C4}
&
\textbf{Wiki}
&
\textbf{C4}
&
\textbf{Wiki}
\\
\midrule

\(\textcolor{Red}{\times}\)
&
\(\textcolor{Red}{\times}\)
&
\(\textcolor{Red}{\times}\)
&
128.77 & 107.92
&
20.56 & 16.36
\\

\(\textcolor{PineGreen}{\checkmark}\)
&
\(\textcolor{Red}{\times}\)
&
\(\textcolor{Red}{\times}\)
&
115.69 & 98.18
&
29.12 & 26.24
\\

\(\textcolor{Red}{\times}\)
&
\(\textcolor{PineGreen}{\checkmark}\)
&
\(\textcolor{Red}{\times}\)
&
119.14 & 99.45
&
19.97 & 15.66
\\

\(\textcolor{Red}{\times}\)
&
\(\textcolor{Red}{\times}\)
&
\(\textcolor{PineGreen}{\checkmark}\)
&
\underline{86.00} & \underline{73.07}
&
19.91 & 16.17
\\

\(\textcolor{PineGreen}{\checkmark}\)
&
\(\textcolor{PineGreen}{\checkmark}\)
&
\(\textcolor{Red}{\times}\)
&
109.53 & 90.58
&
\underline{18.94} & \underline{15.20}
\\

\midrule
\rowcolor{gray!18}
\(\textcolor{PineGreen}{\checkmark}\)
&
\(\textcolor{PineGreen}{\checkmark}\)
&
\(\textcolor{PineGreen}{\checkmark}\)
&
\textbf{64.33} & \textbf{48.11}
&
\textbf{17.74} & \textbf{14.62}
\\

\bottomrule
\end{tabular}
}
}

\vspace{-0.1cm}

\end{wraptable}

Table \ref{tab:true_target_geometry_ablation} ablates true-target reconstruction $\mathcal{L}_{Q}$, the geometry-preservation constraint, and the geometry-guided mask selection. 
 On Qwen2.5-1.5B, replacing the pseudo target with $\mathcal{L}_{Q}$ reduces perplexity from 128.77 to 115.69 on C4 and from 107.92 to 98.18 on WikiText2. In contrast, on LLaMA-2-7B, the same change substantially degrades performance, increasing perplexity from 20.56 to 29.12 and from 16.36 to 26.24, respectively. This result shows that output reconstruction alone does not reliably preserve important representation structure, whereas the geometry constraint stabilizes the reconstruction update across both architectures. Combined with $\mathcal{L}_{Q}$, it improves the baseline by 19.24 and 17.34 perplexity points on Qwen2.5-1.5B, and by 1.62 and 1.16 points on LLaMA-2-7B, over C4 and WikiText2, respectively. 
 Overall, the complete method improves the corresponding baseline by 64.44 and 59.81 points on Qwen2.5-1.5B, and by 2.82 and 1.74 points on LLaMA-2-7B.

\noindent
\textbf{Inference overhead.}
Our method introduces no additional inference-time computation relative to ARB-RC. Both the geometry-preserving optimization and the geometry-guided mask selection are performed only during offline quantization, while the resulting quantized model retains the same inference representation and forward-pass computation as ARB-RC. We further report measured prefill and decoding throughput in Table~\ref{tab:throughput}, confirming comparable inference efficiency.

\section{Conclusion}
In this work, we revisit output-driven optimization for 1-bit post-training quantization of large language models. We show that conventional output reconstruction can distort representation geometry through anisotropic optimization, and address this by formulating quantization as constrained reconstruction with geometry preservation. Experiments across multiple LLM families and benchmarks show consistent improvements in perplexity and downstream accuracy over strong 1-bit PTQ baselines.





\subsection*{AI use statement}

In this work, we used generative AI tools to assist with aspects of implementation and language editing, including improving the clarity and presentation of the manuscript and suggesting improvements to scientific visualizations and figure presentation. All AI-assisted code was reviewed and tested by the authors, and all AI-assisted text and visualization suggestions were reviewed and revised to ensure that they accurately reflect the underlying methods and experimental results. We did not rely on generative AI to generate experimental results or make final methodological decisions. We take responsibility for the final content of this work, including text, claims, code, and artifacts produced with the aid of generative AI.

\bibliography{iclr2027_conference}
\bibliographystyle{iclr2027_conference}
\newpage
\appendix
\crefalias{section}{appendix}

\section{Impact Of Token Geometry Preservation On Attention Mechanism}
\begin{figure*}[!h]
    \centering
    \includegraphics[width=1\textwidth]{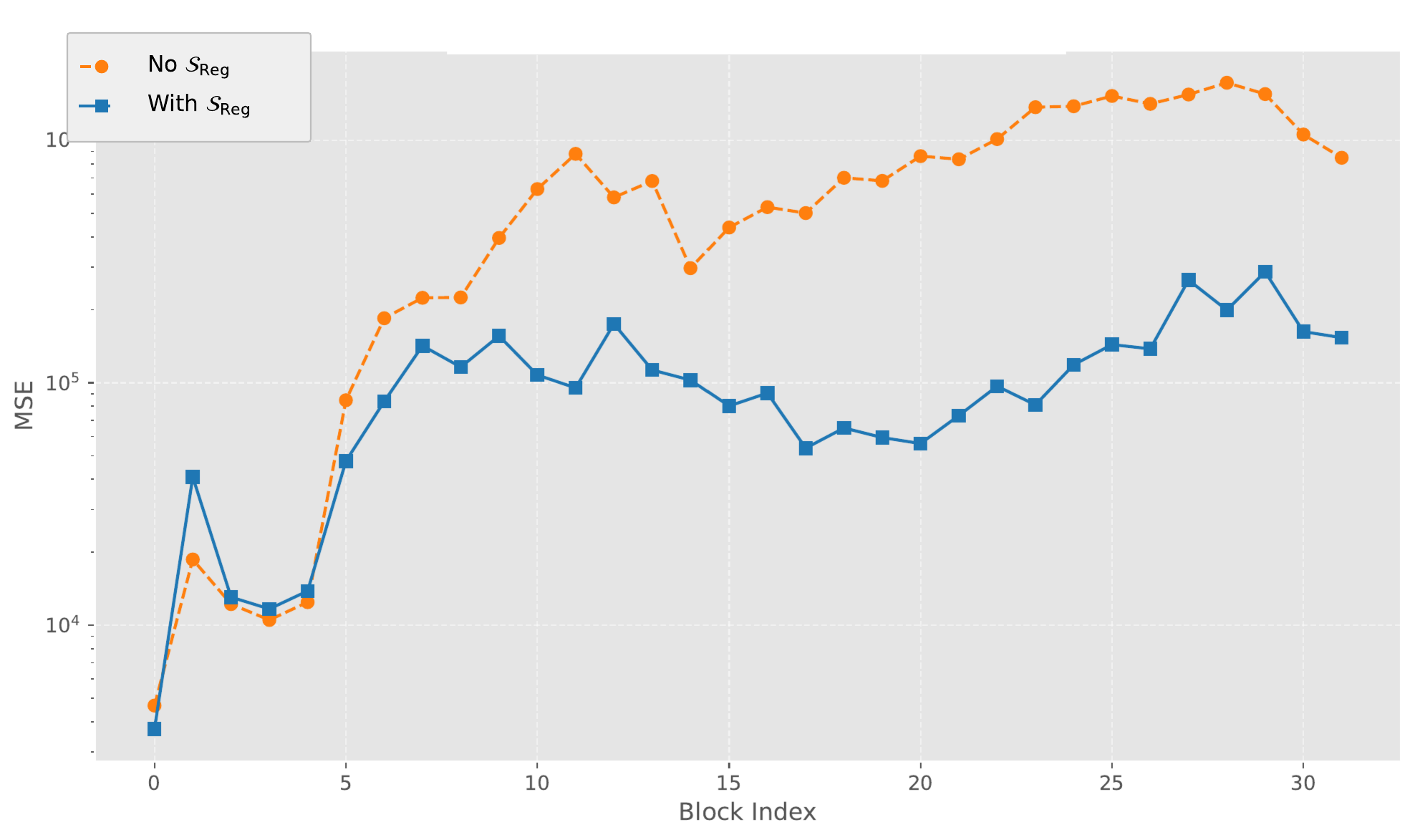}
    \caption{The difference between the token similarity matrix between full-precision model and quantized model, with and without $\mathcal{S}_{Reg}$ constraint, demonstrating the effectiveness of the proposed method.}
\label{fig:geometry_preservation}
    \vspace{-0.1cm}
\end{figure*}

 \paragraph{Effect on true geometry-preserving target.}
 \label{sec:true_geometry_target}
As shown in Fig.~\ref{fig:geometry_preservation}, unconstrained output
optimization progressively distorts the normalized token-similarity structure,
whereas our geometry-constrained updates substantially preserve it across layers.
This confirms that the tractable surrogate in $\mathcal{S}_{Reg}$
effectively protects the normalized geometry.

\begin{figure*}[!h]
    \centering
    \includegraphics[width=1\textwidth]{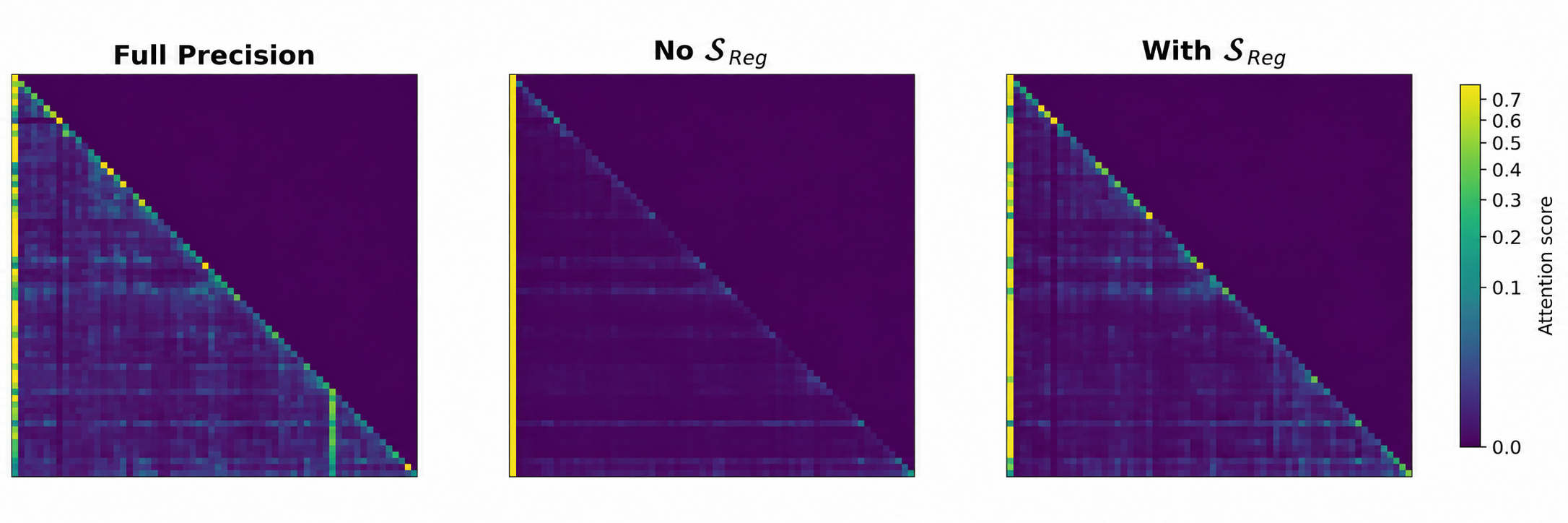}
    \caption{Comparison of attention scores with and without the token-geometry preservation constraint against the full-precision model. Attention scores produced with the $\mathcal{S}_{Reg}$ constraint more closely match those of the pretrained model than those obtained without $\mathcal{S}_{Reg}$ constraint, demonstrating the effectiveness of the proposed method.}
\label{fig:attention_matrix_with_AMP}
    \vspace{-0.1cm}
\end{figure*}

 \paragraph{Effect of geometry-preserving projection.}
 As complementary qualitative evidence, we inspect whether geometry-preserving updates also better retain full-precision attention patterns. Figure~\ref{fig:attention_matrix_with_AMP} qualitatively compares the attention-score patterns produced with and without the geometry-preserving projection. Although the proposed constraint is defined on token-output geometry rather than directly on the attention matrix, the constrained update produces attention patterns that more closely resemble those of the full-precision model. 

\section{Analytical Solutions for Quantization Variables and Geometry Gradient}
\label{app:analytical_solution}

We provide the detailed derivations of the analytical closed-form solutions for the quantization variables $\alpha_r$, $\alpha_c$, and $B$.
\paragraph{Column scales.}
For fixed $\alpha_r$ and $B$, setting
$\nabla_{\alpha_c}\mathcal L_Q=0$ gives
\begin{equation}
\label{eq:alpha_c_closed_form}
    \alpha_c^\star
    =
    \frac{
        \operatorname{diag}
        \left(
            B^\top
            \operatorname{Diag}(\alpha_r)
            SW
        \right)
    }{
        \operatorname{diag}
        \left(
            B^\top
            \operatorname{Diag}(\alpha_r)
            \widehat{S}
            \operatorname{Diag}(\alpha_r)
            B
        \right)
        +
        \varepsilon
    },
\end{equation}
where \(\operatorname{diag}(.)\) denotes the diagonal vector of a matrix, and the division is element-wise. The corresponding update is $\Delta_{\alpha_c}^Q
    =
    \alpha_c^\star-\alpha_c.$

\paragraph{Row scales.}
For fixed $\alpha_c$ and $B$, define $K = \operatorname{Diag}(\alpha_c)^2,
R=BKB^\top.$ The stationary condition for $\alpha_r$ is
\begin{equation}
\label{eq:alpha_r_system}
    (\widehat{S}\odot R)\alpha_r
    =
    \operatorname{diag}
    \left(
        SW
        \operatorname{Diag}(\alpha_c)
        B^\top
    \right).
\end{equation}
The closed-form solution is
\begin{equation}
\label{eq:alpha_r_closed_form}
    \alpha_r^\star
    =
    (\widehat{S}\odot R)^\dagger
    \operatorname{diag}
    \left(
        SW
        \operatorname{Diag}(\alpha_c)
        B^\top
    \right),
\end{equation}
where $(\cdot)^\dagger$ denotes the Moore--Penrose pseudoinverse.
In implementation, we solve
\cref{eq:alpha_r_system} using a numerically stable least-squares solver.
The update is $\Delta_{\alpha_r}^Q
    =
    \alpha_r^\star-\alpha_r.$

\paragraph{Binary rows.}
For fixed $\alpha_r$ and $\alpha_c$, define $N = \operatorname{Diag}(\alpha_r) \widehat{S} \operatorname{Diag}(\alpha_r)$, $K =\operatorname{Diag}(\alpha_c)^2$, $P = \operatorname{Diag}(\alpha_r)SW \operatorname{Diag}(\alpha_c)$. 
We can then derive the closed-form solution as:
\begin{equation}
\label{eq:binary_closed_form}
    B_{i,:}^\star= \mathrm{sign}(P -  [N - \mathrm{diag}(\mathrm{Diag}(N))] B K )_{i,:}.
\end{equation}
The binary update is therefore $\Delta_{B_{i,:}}^Q = B_{i,:}^\star-B_{i,:}.$ Detailed derivations of
Eqs.~\ref{eq:alpha_c_closed_form}, \ref{eq:alpha_r_closed_form}, and \ref{eq:binary_closed_form}
are provided in \cref{sec:derivation_closedform}.

\subsection{Geometry Gradient and Diagonal Curvature}
\label{subsec:geometry_derivatives}

Let $G_{\widehat W} = \nabla_{\widehat W}\mathcal S_{Reg}(\widehat W).$ Applying the chain rule to the parameterization in
\cref{eq:quantized_weight} gives
\begin{equation}
\label{eq:block_gradients}
\begin{aligned}
    g_{\alpha_r}
    &=
    \operatorname{diag}
    \left(
        G_{\widehat W}
        \operatorname{Diag}(\alpha_c)
        B^\top
    \right),\\
    g_{\alpha_c}
    &=
    \operatorname{diag}
    \left(
        B^\top
        \operatorname{Diag}(\alpha_r)
        G_{\widehat W}
    \right),\\
    g_B
    &=
    \operatorname{Diag}(\alpha_r)
    G_{\widehat W}
    \operatorname{Diag}(\alpha_c).
\end{aligned}
\end{equation}
For a binary-row update, $g_{B_{i,:}}$ is the $i$-th row of $g_B$. The corresponding diagonal curvature vectors are
\begin{equation}
\label{eq:regularizer_hessian_diagonals}
\begin{aligned}
    h_{\alpha_r}
    &=
    \operatorname{diag}(M)
    \odot
    \operatorname{diag}
    \left(
        B D_c^2 B^\top
    \right),\\
    h_{\alpha_c}
    &=
    \operatorname{diag}
    \left(
        B^\top D_r M D_r B
    \right),\\
    h_B
    &=
    \left(
        \alpha_r^{\odot 2}
        \odot
        \operatorname{diag}(M)
    \right)
    \left(
        \alpha_c^{\odot 2}
    \right)^\top.
\end{aligned}
\end{equation}
The gradient and diagonal curvature are
substituted directly into \cref{eq:geometry_mask}.

\section{Proof of Proposition~\ref{prop:geometry_preservation}}
\label{app:amp_proof}

\begin{proof}
Since $\mathcal{S}_{\mathrm{Reg}}$ is quadratic in each conditional
variable block $\phi$, its change around the current point is
\[
\mathcal{S}_{\mathrm{Reg}}(\phi+\Delta)
-
\mathcal{S}_{\mathrm{Reg}}(\phi)
=
g_{\phi}^{\top}\Delta
+
\frac{1}{2}\Delta^{\top}H_{\phi}\Delta.
\]
As introduced in Eq.~\ref{eq:geometry_taylor}, we approximate this
coupled quadratic change using the separable diagonal-curvature model
\[
\widetilde{\Delta\mathcal{S}}_{\mathrm{Reg}}(\Delta)
=
\sum_j
\left[
(g_{\phi})_j\Delta_j
+
\frac{\beta}{2}(h_{\phi})_j\Delta_j^2
\right].
\]
For every retained coordinate,
$\Delta_j=(\Delta_{\phi}^{Q})_j$, and the corresponding term is
nonnegative by construction of $M_{\phi}$. For every rejected coordinate,
$\Delta_j=0$, and the corresponding term is zero. Therefore, every term
in the sum is nonnegative, and hence
\[
\widetilde{\Delta\mathcal{S}}_{\mathrm{Reg}}
(\widetilde{\Delta}_{\phi})
=
\sum_j
\left[
(g_{\phi})_j\widetilde{\Delta}_{\phi,j}
+
\frac{\beta}{2}(h_{\phi})_j
\widetilde{\Delta}_{\phi,j}^{2}
\right]
\geq 0.
\]
Thus, $\widetilde{\Delta}_{\phi}$ satisfies the separable local
geometry constraint in Eq.~\ref{eq:geometry_taylor}.
\end{proof}


\section{Derivation of closed-forms}
\label{sec:derivation_closedform}
\paragraph{Derivation for $\alpha_c$ } 
The gradient of the $\mathcal{L}_{Q}$ w.r.t $\alpha_c$ is: 

\begin{equation}
\label{eq:derivation_alpha_c_1}
\begin{aligned}
\frac{\partial \mathcal{L}_{Q}}{\partial \mathrm{diag}(\alpha_c)}
&= \frac{\partial \mathcal{L}_{Q}}
        {\partial (X W - \widehat{X} \widehat{W})}
   \frac{\partial (X W - \widehat{X} \widehat{W})}
        {\partial \alpha_c} \\
&= -2 B^{\top} \mathrm{diag}(\alpha_r)\widehat{X}^{\top} \times (X W - \widehat{X} \widehat{W}) \\
&= -2 B^{\top} \mathrm{diag}(\alpha_r)\widehat{X}^{\top} \times
(X W - \widehat{X}\mathrm{diag}(\alpha_r)B\mathrm{diag}(\alpha_c)).
\end{aligned}
\end{equation}

Setting the diagonal of this gradient to 0, we have:
\begin{equation}
\label{eq:derivation_alpha_c_2}
\begin{aligned}
   &\mathrm{Diag}({B^{\top} \mathrm{diag}(\alpha_r)\widehat{S} \mathrm{diag}(\alpha_r)B}) \odot \alpha_c = \mathrm{Diag}(B^{\top} \mathrm{diag}(\alpha_r) S W) \\
   &\implies \alpha_c = \frac{\mathrm{Diag}({B}^{T}\mathrm{diag}(\alpha_r) S W )}{\mathrm{Diag}({B}^{T}\mathrm{diag}(\alpha_r) \widehat{S}\mathrm{diag}(\alpha_r) B )}.
\end{aligned}
\end{equation}

\paragraph{Derivation for $\alpha_r$}
The gradient of the objective w.r.t $\alpha_r$ is:
\begin{equation}
\label{eq:derivation_alpha_r_1}
\begin{aligned}
   \frac{\partial \mathcal{L}}{\partial \mathrm{diag}(\alpha_r)} & = \frac{\partial \mathcal{L}}{\partial (X W - \widehat{X} \widehat{W})} \frac{\partial (X W - \widehat{X} \widehat{W})}{\partial \alpha_r} \\
   &= - 2\widehat{X}^{\top} (X W - \widehat{X} \widehat{W}) \mathrm{diag}(\alpha_c) B^{\top}  \\
   &= - 2\widehat{X}^{\top} (X W - \widehat{X} \mathrm{diag}(\alpha_r)B\mathrm{diag}(\alpha_c)) \mathrm{diag}(\alpha_c) B^{\top} \\
\end{aligned}
\end{equation}
With $\widehat{S} = \widehat{X}^{\top} \widehat{X}$, setting the diagonal of this gradient to 0, we have: 
\begin{equation}
\label{eq:derivation_alpha_r_2}
\begin{aligned}
   &\quad\mathrm{Diag}(\widehat{S} \mathrm{diag}(\alpha_r)B \mathrm{diag}(\alpha_c \odot \alpha_c) B^{\top}) = \mathrm{Diag}( S W \mathrm{diag}(\alpha_c) B^{\top} ).
\end{aligned}
\end{equation}
Let $C=B \mathrm{diag}(\alpha_c \odot \alpha_c) B^{\top}$, then we have:
\begin{equation}
\begin{aligned}
     &\mathrm{Diag}(\widehat{S} \mathrm{diag}(\alpha_r)  C  )  =  \mathrm{Diag}( S W \mathrm{diag}(\alpha_c) B^{\top} ) \\
   & \implies  (\widehat{S} \odot C) \alpha_r   =   \mathrm{Diag}( S W \mathrm{diag}(\alpha_c) B^{\top} ) \\
   & \implies  \alpha_r   = \big( \widehat{S} \odot C \big)^{-1} 
\, \mathrm{Diag}\!\left( S W \mathrm{diag}(\alpha_c) {B}^{\top} \right)
\end{aligned}
\end{equation}

\paragraph{Derivation for $B$}
Explanding the objective loss $\mathcal{L}(.)$ out we have:

\begin{equation}
\label{eq:derivation_B}
\begin{aligned}
    & \operatorname{Tr}\left[(X W - \widehat{X} \widehat{W})(X W - \widehat{X} \widehat{W})^{T}\right] \\
   &= \operatorname{Tr}\left[\widehat{X}\widehat{W} \widehat{W}^{\top}\widehat{X}^{\top} - 2X W  \widehat{W}^{\top}\widehat{X}^{\top}  + \text{const}\right] \\
   &\propto \operatorname{Tr}\left[\widehat{X}^{\top}\widehat{X}\widehat{W} \widehat{W}^{\top}\right]  - 2 \operatorname{Tr}\left[ \widehat{X}^{\top} X W\widehat{W}^{\top}\right]    \\
   &= \operatorname{Tr}\left[\widehat{S}\widehat{W} \widehat{W}^{\top}\right]  - 2 \operatorname{Tr}\left[ S W \widehat{W}^{\top}\right]    \\
   &= \operatorname{Tr}\left[\mathrm{diag}(\alpha_r)\widehat{S}\mathrm{diag}(\alpha_r)B \mathrm{diag}(\alpha_c \odot \alpha_c)B^{\top}\right]
   - 2 \operatorname{Tr}\left[\mathrm{diag}(\alpha_r)S W \mathrm{diag}(\alpha_c) B^{\top}\right]   \\
   &= \operatorname{Tr}\left[N B  K  B^{\top}\right]   - 2 \operatorname{Tr}\left[ P  B^{\top}\right]   \\
\end{aligned}
\end{equation}

with $N = \mathrm{diag}(\alpha_r)\widehat{S}\mathrm{diag}(\alpha_r)$, $K = \mathrm{diag}(\alpha_c \odot \alpha_c)$, $P = \mathrm{diag}(\alpha_r)S W \mathrm{diag}(\alpha_c)$. 

We denote $\widetilde{B}_{-i}$ and $\widetilde{N}_{-i}$  respectively as the matrix $B$ and $N$ exclude the row $i$. Similarly, $\widetilde{N}_{i,-j}$ denotes the row i of matrix $N$ excluded the $j$ element. Given a row $i$ of $B$, keeping the other row of $B$ as constant, the loss can be expanded as a function of $B_{i,:}$ as:

\begin{equation}
\label{eq:derivation_B_2}
\begin{aligned}
   \mathcal{L}_{MSE}(X,L)
   &\propto \operatorname{Tr}\left[N B K B^{\top}\right]
   - 2 \operatorname{Tr}\left[P B^{\top}\right] \\
   &= \sum_{j} \left(
   N_{j,:} B K B_{j,:}^{\top}
   - 2 P_{j,:}B_{j,:}^{\top}
   \right) \\
   &\propto
   N_{i,i} B_{i,:} K B_{i,:}^{\top}
   + 2\widetilde{N}_{i,-i}\widetilde{B}_{-i}K B_{i,:}^{\top}
   - 2P_{i,:}B_{i,:}^{\top} \\
   &\propto
   \left(
   \widetilde{N}_{i,-i}\widetilde{B}_{-i}K
   - P_{i,:}
   \right)B_{i,:}^{\top},
\end{aligned}
\end{equation}
Since $B_{i,:}$ is a binary vector, it has a closed-form solution:

\begin{equation}
\label{eq:derivation_B_3}
\begin{aligned}
   B_{i,:}
   = \mathrm{sign}\left(
   P_{i,:}
   - \widetilde{N}_{i,-i}\widetilde{B}_{-i}K
   \right).
\end{aligned}
\end{equation}

The more compacted form of the whole $B$ using this closed-form for each row can be computed as:
\begin{equation}
\label{eq:derivation_B_4}
\begin{aligned}
   B= \mathrm{sign}(P -  [N - \mathrm{diag}(\mathrm{Diag}(N))] B K ).
\end{aligned}
\end{equation}

\section{Additional results}
\label{subsec:addtional_result}

\paragraph{Quantization overhead.}
\label{sec:offline_overhead}
We report the quantization time of our method compared with ARB-X and ARB-RC~\cite{Li2024ARBLLMAR} across different architectures. Our method incurs additional offline quantization cost relative to ARB-RC, while remaining comparable to ARB-X, it is faster than ARB-X on both LLaMA models and slightly slower on OPT-6.7B. In terms of memory usage, for the Llama-2-7B model, our method requires a peak GPU memory of 12.1\,GB, compared with 11.4\,GB for ARB-X and 10.8\,GB for ARB-RC. As quantization is performed only once offline, this modest computational and memory overhead is modest in practice.

\begin{table}[t]
    \caption{Quantization time comparison between 1-bit LLM methods and ours across different models.}
    \label{tab:time_comparison}
    \centering
    
    \begin{tabular}{lccc}
        \toprule
        \textbf{Model} & \textbf{ARB-X} & \textbf{ARB-RC} & \textbf{Ours} \\
        \midrule
        \textbf{OPT-6.7B}    & 87m  & 65m  & 93m  \\
        \textbf{Llama-2-7B}  & 91m  & 54m  & 76m  \\
        \textbf{Llama-2-13B} & 147m & 100m & 119m \\
        \bottomrule
    \end{tabular}

\end{table}

\subsection{Additional results on the efficiency of the model}

We provide  the throughput improvements of the Prefill and Decode passes (single batch with 2048 tokens) for quantized models, relative to their FP16 counterparts. The base model used is LLaMA-2-7B, with inferences conducted on a Intel Xeon Gold 6548Y+ CPU. Overall, our method achieves at least 4.4x speed up in both decoding and prefill compared to the FP counterpart, and comparable to the ARB-RC. 




\begin{table}[t]
    \caption{Inference throughput (tokens/s) on the Intel Xeon Gold 6548Y+ CPU \cite{Wang2023BitNetS1}. Higher is better.}
    \label{tab:throughput}
    \centering
    \begin{tabular}{l|cc}
        \toprule
        \textbf{Model} & \textbf{Prefill} & \textbf{Decode}  \\
        \midrule
        \textbf{FP16}    & 2.83 & 0.87  \\
        \textbf{ARB-RC}  & 11.44  & 14.15 \\
        \textbf{Ours} & 12.42 & 14.52 \\
        \bottomrule
    \end{tabular}

\end{table}

\subsection{Additional implementation details}

It is widely known that block-level loss are more important than layer-wise loss, due to cross-layer dependency across layers within blocks \citep{li2021brecq}. However, naïvely applying output-driven quantization to all layers within a block does not necessarily guarantee minimization of the block-level loss. To ensure stability, we therefore adopt a conservative design choice and restrict output alignment to only the last fully connected layer of each block, which has a direct and dominant influence on the block loss. The remaining layers within each block are quantized using weight-alignment objective of ARB-RC \cite{Li2024ARBLLMAR}.

\paragraph{Ablation study for the size of the calibration set.}
We provide additional ablation study about the effect of our method when the size of the calibration set varies. The results are provided in Table \ref{tab:calibration_size_opt}.  The results demonstrate that our method are effective both when the number of calibration samples is small, or when we use more data. 
\begin{table}[t]
\centering
\caption{Ablation study on the calibration set size for Qwen-2-1.5B.}
\label{tab:calibration_size_opt}
\begin{tabular}{c|c|cc}
    \toprule
    \multirow{2}{*}{\textbf{Size}} 
    & \multirow{2}{*}{\textbf{Method}} 
    & \multicolumn{2}{c}{\textbf{PPL} $(\downarrow)$} \\ 
    \cmidrule(r){3-4}
     &  & \textbf{C4} & \textbf{WikiText2}  \\
    \midrule
    \multirow{2}{*}{32}  
      & ARB-RC & 109.45 & 95.08 \\
      & Ours   & \textbf{83.44} & \textbf{76.56} \\
    \cmidrule(r){1-4}
    \multirow{2}{*}{128} 
      & ARB-RC & 109.83 & 89.59 \\
      & Ours   & \textbf{74.83} & \textbf{68.45} \\
    \cmidrule(r){1-4}
    \multirow{2}{*}{512} 
      & ARB-RC & 127.84 & 89.13 \\
      & Ours   & \textbf{71.27} & \textbf{54.66} \\
    \cmidrule(r){1-4}
    \multirow{2}{*}{1024} 
      & ARB-RC & 117.49 & 92.29 \\
      & Ours   & \textbf{67.80} & \textbf{60.49} \\
    \bottomrule
\end{tabular}

\end{table}

\paragraph{Ablation study for the value of the $\beta$.}
We provide additional ablation study about the value of $\beta$. The results are provided in Table \ref{tab:ablation_beta}. In practice we set the value of $\beta=1$ for simplicity. We use a single fixed seed for this ablation study.

\begin{table}[H]
\centering
\caption{Ablation study for the hyper-parameter $\beta$.}
\label{tab:ablation_beta}
\renewcommand\tabcolsep{6pt}

\begin{tabular}{l|c|cc}
    \toprule[1.2pt]
    \multirow{2}{*}{\textbf{Model}} & \multirow{2}{*}{$\beta$} & \multicolumn{2}{c}{\textbf{PPL} $(\downarrow)$} \\ 
    \cmidrule(r){3-4}
     &  & \textbf{C4} & \textbf{WikiText2}  \\
    \midrule
    \multirow{3}{*}{\textbf{OPT-6.7B}} & 0  & 16.24 & 14.56 \\
              & 0.5 & 16.21 & 14.53 \\
              & 1 & 16.18 & 14.59 \\
    \bottomrule[1.2pt]
\end{tabular}

\end{table}

\section{Proposed learning algorithm}
\label{app:learning_algorithm}

\begin{algorithm*}[th]
\caption{Our-RC with Second-Order Relational Projection}
\label{alg:our-rc}
\begin{algorithmic}[1]
\Procedure{Our-RC}{$W,S,\widehat S,T,k,\beta$}
    \LComment{$W\in\mathbb R^{d_{\mathrm{in}}\times d_{\mathrm{out}}}$:
    full-precision weight}
    \LComment{$S=\widehat X^\top X\in
    \mathbb R^{d_{\mathrm{in}}\times d_{\mathrm{in}}}$}
    \LComment{$\widehat S=\widehat X^\top\widehat X\in
    \mathbb R^{d_{\mathrm{in}}\times d_{\mathrm{in}}}$}
    \LComment{$T$: refinement rounds; $k$: column-scale update interval}
    \LComment{$\beta$: diagonal curvature multiplier}
    \State Initialize $\alpha_r,\alpha_c,B$
    \State $\widehat W\gets
    \operatorname{Diag}(\alpha_r)B\operatorname{Diag}(\alpha_c)$
    \State $M_G\gets SWW^\top S^\top$

    \For{$\mathrm{iter}=1,\ldots,T$}

        \State $\alpha_r^\star\gets$
        \Call{Refine-$\alpha_r$}
        {$S,\widehat S,W,B,\alpha_c$}
        \State $\mathcal M^r\gets$
        \Call{SO-$\alpha_r$}
        {$M_G,B,\alpha_r,\alpha_c,\alpha_r^\star,\beta$}
        \State $\widetilde\alpha_r\gets
        (1-\mathcal M^r)\odot\alpha_r+
        \mathcal M^r\odot\alpha_r^\star$
        \State $\widehat W_{\mathrm{new}}\gets
        \operatorname{Diag}(\widetilde\alpha_r)
        B\operatorname{Diag}(\alpha_c)$
        \If{\Call{Accept}
        {$\widehat W,\widehat W_{\mathrm{new}},W,S,\widehat S,M_G$}}
            \State $\alpha_r\gets\widetilde\alpha_r$
            \State $\widehat W\gets\widehat W_{\mathrm{new}}$
        \EndIf

        \If{$(\mathrm{iter}+1)\bmod k=0$}
            \State $\alpha_c^\star\gets$
            \Call{Refine-$\alpha_c$}
            {$S,\widehat S,W,B,\alpha_r$}
            \State $\mathcal M^c\gets$
            \Call{SO-$\alpha_c$}
            {$M_G,B,\alpha_r,\alpha_c,\alpha_c^\star,\beta$}
            \State $\widetilde\alpha_c\gets
            (1-\mathcal M^c)\odot\alpha_c+
            \mathcal M^c\odot\alpha_c^\star$
            \State $\widehat W_{\mathrm{new}}\gets
            \operatorname{Diag}(\alpha_r)
            B\operatorname{Diag}(\widetilde\alpha_c)$
            \If{\Call{Accept}
            {$\widehat W,\widehat W_{\mathrm{new}},W,S,\widehat S,M_G$}}
                \State $\alpha_c\gets\widetilde\alpha_c$
                \State $\widehat W\gets\widehat W_{\mathrm{new}}$
            \EndIf
        \EndIf

        \For{$q=1,2$}
            \State $(B^\star,i)\gets$
            \Call{Refine-B}
            {$S,\widehat S,W,B,\alpha_r,\alpha_c$}
            \State $\mathcal M^B_{i,:}\gets$
            \Call{SO-B}
            {$M_G,B,B^\star,i,\alpha_r,\alpha_c,\beta$}
            \State $\widetilde B\gets B$
            \State $\widetilde B_{i,:}\gets
            (1-\mathcal M^B_{i,:})\odot B_{i,:}
            +
            \mathcal M^B_{i,:}\odot B^\star_{i,:}$
            \State $\widehat W_{\mathrm{new}}\gets
            \operatorname{Diag}(\alpha_r)
            \widetilde B
            \operatorname{Diag}(\alpha_c)$
            \If{\Call{Accept}
            {$\widehat W,\widehat W_{\mathrm{new}},W,S,\widehat S,M_G$}}
                \State $B\gets\widetilde B$
                \State $\widehat W\gets\widehat W_{\mathrm{new}}$
            \EndIf
        \EndFor
    \EndFor

    \State $\widehat W\gets
    \operatorname{Diag}(\alpha_r)
    B
    \operatorname{Diag}(\alpha_c)$
    \State \Return $\widehat W$
\EndProcedure
\end{algorithmic}
\end{algorithm*}

    \begin{algorithm*}[th]
\caption{Auxiliary Functions}
\label{alg:auxiliary}
\begin{multicols}{2}
\begin{algorithmic}[1]

\Procedure{Refine-$\alpha_r$}
{$S,\widehat S,W,B,\alpha_c$}
    \State $K\gets
    \operatorname{Diag}(\alpha_c\odot\alpha_c)$
    \State $R\gets BKB^\top$
    \State $H_r^Q\gets\widehat S\odot R$
    \State $b_r^Q\gets
    \operatorname{diag}
    \left(
        SW\operatorname{Diag}(\alpha_c)B^\top
    \right)$
    \State $\alpha_r^\star\gets$
    \Call{Solve}{$H_r^Q,b_r^Q$}
    \State \Return $\alpha_r^\star$
\EndProcedure

\Statex

\Procedure{Refine-$\alpha_c$}
{$S,\widehat S,W,B,\alpha_r$}
    \State $D_r\gets\operatorname{Diag}(\alpha_r)$
    \State $\mathrm{num}\gets
    \operatorname{diag}
    \left(
        B^\top D_rSW
    \right)$
    \State $\mathrm{den}\gets
    \operatorname{diag}
    \left(
        B^\top D_r\widehat S D_rB
    \right)$
    \State $\alpha_c^\star\gets
    \mathrm{num}/(\mathrm{den}+\varepsilon)$
    \State \Return $\alpha_c^\star$
\EndProcedure

\Statex

\Procedure{Refine-B}
{$S,\widehat S,W,B,\alpha_r,\alpha_c$}
    \State $D_r\gets\operatorname{Diag}(\alpha_r)$
    \State $D_c\gets\operatorname{Diag}(\alpha_c)$
    \State $N\gets D_r\widehat S D_r$
    \State $K\gets D_c^2$
    \State $P\gets D_rSWD_c$
    \State $N_{\mathrm{off}}\gets
    N-\operatorname{Diag}(\operatorname{diag}(N))$
    \State $Z\gets P-N_{\mathrm{off}}BK$
    \State $B^\star\gets\operatorname{sign}(Z)$
    \State $\mathrm{gain}_j\gets
    \sum_{\ell}
    \left(
        |Z_{j\ell}|-B_{j\ell}Z_{j\ell}
    \right)$
    \State $i\gets\operatorname*{argmax}_j\mathrm{gain}_j$
    \State \Return $B^\star,i$
\EndProcedure

\Statex

\Procedure{SO-$\alpha_r$}
{$M_G,B,\alpha_r,\alpha_c,\alpha_r^\star,\beta$}
    \State $D_c\gets\operatorname{Diag}(\alpha_c)$
    \State $\widehat W\gets
    \operatorname{Diag}(\alpha_r)BD_c$
    \State $\Delta_r\gets\alpha_r^\star-\alpha_r$
    \State $U\gets BD_c$
    \State $g_r\gets
    \operatorname{diag}
    \left(
        M_G\widehat W D_cB^\top
    \right)$
    \State $h_r\gets
    \operatorname{diag}(M_G)
    \odot
    \operatorname{diag}(UU^\top)$
    \State $\mathcal M^r\gets
    \mathbf 1
    \left[
        g_r\odot\Delta_r+
        \frac{\beta}{2}
        h_r\odot\Delta_r^{\odot2}
        \geq0
    \right]$
    \State \Return $\mathcal M^r$
\EndProcedure

\Statex

\Procedure{SO-$\alpha_c$}
{$M_G,B,\alpha_r,\alpha_c,\alpha_c^\star,\beta$}
    \State $D_r\gets\operatorname{Diag}(\alpha_r)$
    \State $\widehat W\gets
    D_rB\operatorname{Diag}(\alpha_c)$
    \State $\Delta_c\gets\alpha_c^\star-\alpha_c$
    \State $V\gets D_rB$
    \State $g_c\gets
    \operatorname{diag}
    \left(
        B^\top D_rM_G\widehat W
    \right)$
    \State $h_c\gets
    \operatorname{diag}
    \left(
        V^\top M_GV
    \right)$
    \State $\mathcal M^c\gets
    \mathbf 1
    \left[
        g_c\odot\Delta_c+
        \frac{\beta}{2}
        h_c\odot\Delta_c^{\odot2}
        \geq0
    \right]$
    \State \Return $\mathcal M^c$
\EndProcedure

\Statex

\Procedure{SO-B}
{$M_G,B,B^\star,i,\alpha_r,\alpha_c,\beta$}
    \State $D_r\gets\operatorname{Diag}(\alpha_r)$
    \State $D_c\gets\operatorname{Diag}(\alpha_c)$
    \State $\widehat W\gets D_rBD_c$
    \State $\Delta_B\gets B^\star_{i,:}-B_{i,:}$
    \State $g_B\gets
    \left(
        D_rM_G\widehat W D_c
    \right)_{i,:}$
    \State $h_B\gets
    \alpha_{r,i}^2(M_G)_{ii}
    (\alpha_c\odot\alpha_c)$
    \State $\mathcal M^B_{i,:}\gets
    \mathbf 1
    \left[
        g_B\odot\Delta_B+
        \frac{\beta}{2}
        h_B\odot\Delta_B^{\odot2}
        \geq0
    \right]$
    \State \Return $\mathcal M^B_{i,:}$
\EndProcedure

\Statex

\Procedure{Accept}
{$\widehat W_{\mathrm{old}},
  \widehat W_{\mathrm{new}},
  W,S,\widehat S,M_G$}
    \State $\mathcal L_{\mathrm{old}}\gets
    \operatorname{Tr}
    (\widehat W_{\mathrm{old}}^\top
     \widehat S
     \widehat W_{\mathrm{old}})
    -
    2\operatorname{Tr}
    (\widehat W_{\mathrm{old}}^\top SW)$
    \State $\mathcal L_{\mathrm{new}}\gets
    \operatorname{Tr}
    (\widehat W_{\mathrm{new}}^\top
     \widehat S
     \widehat W_{\mathrm{new}})
    -
    2\operatorname{Tr}
    (\widehat W_{\mathrm{new}}^\top SW)$
    \State $\mathcal A_{\mathrm{old}}\gets
    \frac{1}{2}
    \operatorname{Tr}
    (\widehat W_{\mathrm{old}}^\top
     M_G
     \widehat W_{\mathrm{old}})$
    \State $\mathcal A_{\mathrm{new}}\gets
    \frac{1}{2}
    \operatorname{Tr}
    (\widehat W_{\mathrm{new}}^\top
     M_G
     \widehat W_{\mathrm{new}})$
    \State \Return
    $\mathbf 1[
        \mathcal L_{\mathrm{new}}
        \leq\mathcal L_{\mathrm{old}}
        \land
        \mathcal A_{\mathrm{new}}
        \geq\mathcal A_{\mathrm{old}}
    ]$
\EndProcedure

\end{algorithmic}
\end{multicols}
\end{algorithm*}

\clearpage

\end{document}